\documentclass[10pt,journal,compsoc]{IEEEtran}

\usepackage{amsmath,amsfonts,bm}

\def\eqref#1{equation~\ref{#1}}

\def\1{\bm{1}}

\def\vc{{\bm{c}}}

\def\vn{{\bm{n}}}

\def\vx{{\bm{x}}}
\def\vy{{\bm{y}}}
\def\vz{{\bm{z}}}

\def\mG{{\bm{G}}}
\def\mH{{\bm{H}}}
\def\mI{{\bm{I}}}
\def\mJ{{\bm{J}}}
\def\mK{{\bm{K}}}

\def\mM{{\bm{M}}}

\def\mR{{\bm{R}}}

\def\mX{{\bm{X}}}

\DeclareMathAlphabet{\mathsfit}{\encodingdefault}{\sfdefault}{m}{sl}
\SetMathAlphabet{\mathsfit}{bold}{\encodingdefault}{\sfdefault}{bx}{n}

\newcommand{\etal}{{et al. }}

\usepackage{microtype}
\usepackage{graphicx}
\usepackage{booktabs} 
\usepackage{caption}
\usepackage{subcaption}
\usepackage{multirow}
\usepackage{graphics}
\usepackage{amsmath}
\usepackage{hyperref}
\usepackage{amssymb}
\usepackage{url}
\usepackage{graphics}
\usepackage{makecell}

\ifCLASSOPTIONcompsoc
 \usepackage[nocompress]{cite}
\else
 \usepackage{cite}
\fi

\ifCLASSINFOpdf

\else

\fi

\usepackage{xcolor}
\usepackage{ulem}

\begin{document}

\title{Zero-Shot Neural Architecture Search: Challenges, Solutions, and Opportunities}

\author{Guihong Li,~\IEEEmembership{Student Member,~IEEE,}
 Duc Hoang,~\IEEEmembership{Student Member,~IEEE,}\\
 Kartikeya Bhardwaj,~\IEEEmembership{Member,~IEEE,}
 Ming Lin,~\IEEEmembership{Member,~IEEE,}\\
 Zhangyang Wang,~\IEEEmembership{Senior Member,~IEEE,}
 Radu Marculescu,~\IEEEmembership{Fellow,~IEEE}
\IEEEcompsocitemizethanks{\IEEEcompsocthanksitem Guihong Li, Duc Hoang, Zhangyang Wang, and Radu Marculescu are with the Department of Electrical and Computer Engineering, The University
of Texas at Austin, TX, 78712. E-mail: \{lgh, hoangduc, atlaswang, radum\}@utexas.edu
\IEEEcompsocthanksitem Kartikeya Bhardwaj is with Qualcomm AI Research, an initiative of Qualcomm Technologies, Inc., CA, 92121. E-mail: kbhardwa@qti.qualcomm.com
\IEEEcompsocthanksitem Ming Lin is with Amazon, WA, 98004. E-mail: minglamz@amazon.com.
\IEEEcompsocthanksitem Correspondence to Radu Marculescu (radum@utexas.edu).
}

}

\markboth{IEEE TRANSACTIONS ON PATTERN ANALYSIS AND MACHINE INTELLIGENCE}%
{Shell \MakeLowercase{\textit{et al.}}: Bare Demo of IEEEtran.cls for Computer Society Journals}

\IEEEtitleabstractindextext{%
\begin{abstract}
Recently, \textit{zero-shot} (or \textit{training-free}) Neural Architecture Search (NAS) approaches have been proposed to liberate NAS from the expensive training process. The key idea behind zero-shot NAS approaches is to design proxies that can predict the accuracy of some given networks without training the network parameters. The proxies proposed so far are usually inspired by recent progress in theoretical understanding of deep learning and have shown great potential on several datasets and NAS benchmarks. This paper aims to comprehensively review and compare the state-of-the-art (SOTA) zero-shot NAS approaches, with an emphasis on their hardware awareness. To this end, we first review the mainstream zero-shot proxies and discuss their theoretical underpinnings. 
We then compare these zero-shot proxies through large-scale experiments and demonstrate their effectiveness in both hardware-aware and hardware-oblivious NAS scenarios. Finally, we point out several promising ideas to design better proxies. Our source code and the list of related papers are available on \url{https://github.com/SLDGroup/survey-zero-shot-nas}.
\end{abstract}

}

\maketitle

\IEEEdisplaynontitleabstractindextext

\IEEEpeerreviewmaketitle

\IEEEraisesectionheading{\section{Introduction}\label{sec:intro}}
In recent years, deep neural networks have made significant breakthroughs in many applications, such as recommendation systems, image classification, and natural language modeling~\cite{alexnet, vgg, densenet,resnet,dosovitskiy2021an_vit, big_nlp, vaswani2017attention_transfomer}. To automatically design high performance deep networks, \textit{Neural Architecture Search} (NAS) has been proposed during the past decade~\cite{baker_17, quoc_le, lstm, Darts,elsken2019neural}. Specifically, NAS boils down to solving an optimization problem with specific targets (\textit{e.g.,} high classification accuracy) over a set of possible candidate architectures (search space) within a group of computational budgets. Recent breakthroughs in NAS simplify the trial-and-error manual architecture design process and discover new deep network architectures with better performance and efficiency over hand-crafted ones~\cite{Darts,lstm,hyper_nas,real_17,gong2019autogan,xie2018snas,wu2019fbnet,wan2020fbnetv2,li2020random,kandasamy2018neural_opt,yu2019evaluating,liu2017hierarchical,cai2018efficient,luo2018neural,zhang2018graph,zhou2019bayesnas,howard2019searching,yu2020bignas}. Therefore, NAS has attracted significant attention from both academia and industry.

\begin{figure}[tb]
 \centering
 \includegraphics[width=0.5\textwidth]{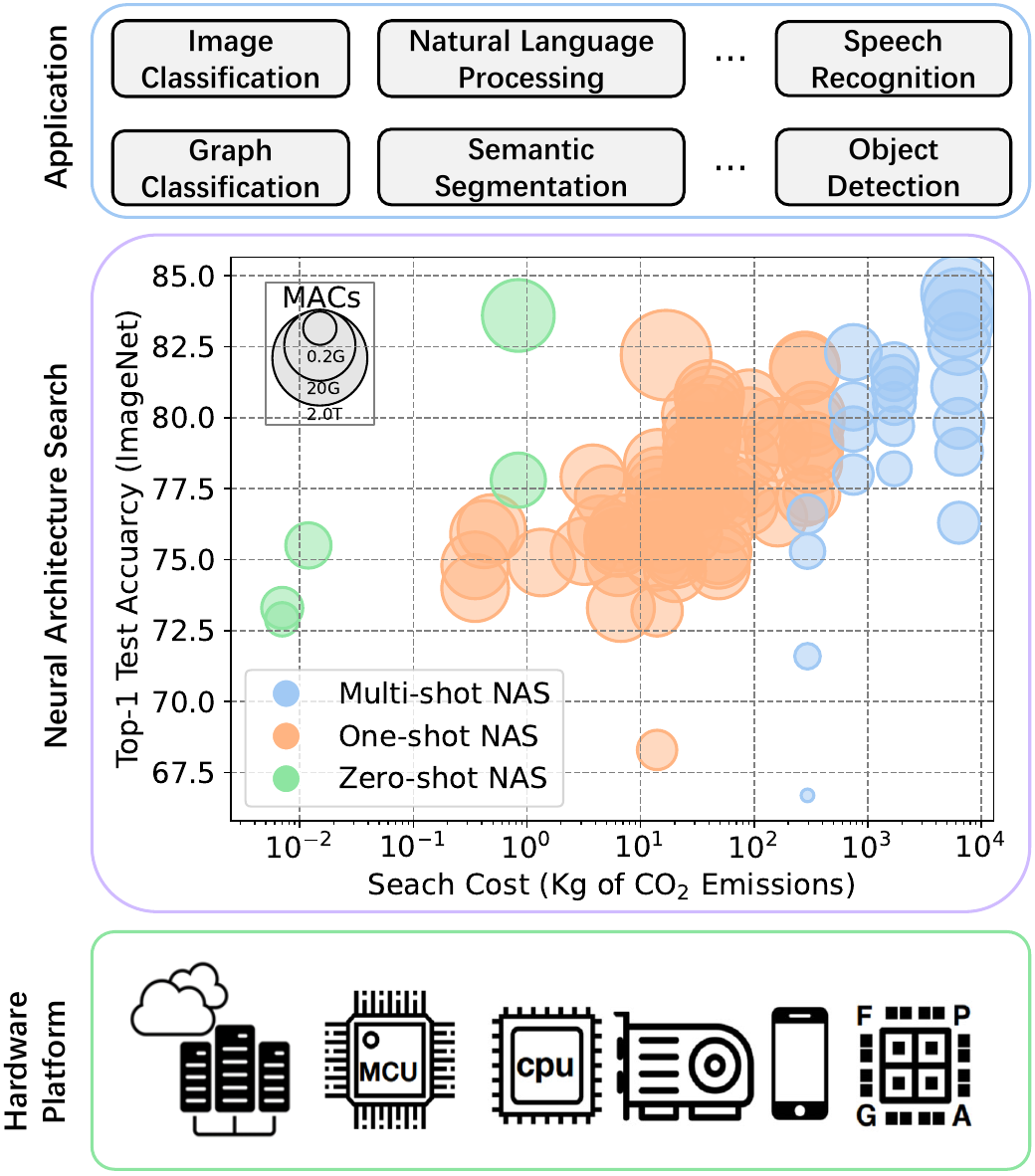}
 \caption{Overview of existing NAS approaches. NAS is designed to search for optimal architectures with both good accuracy and high efficiency on real hardware. (Data collected from~\cite{paper_with_code_nas})}
 \label{fig:nas_compare}
\end{figure}

\begin{figure}[tb]
 \centering
 \includegraphics[width=0.45\textwidth]{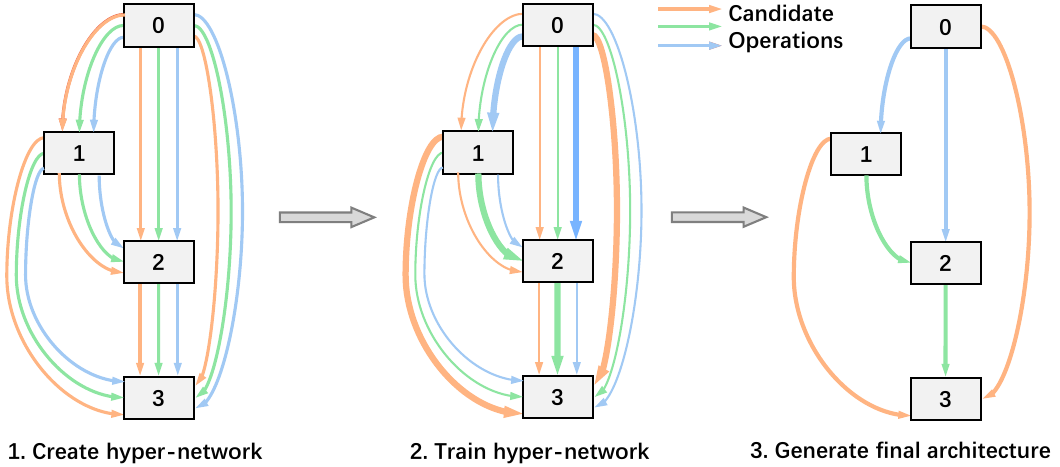}
 \caption{Illustration of differentiable neural architecture search. (1). Merge all candidate operations into a hyper-network with learnable weights for each operation. (2). Train the hyper-network and update the learnable weights for each operation. (3) Generate the final results by selecting the operations with the highest weight values (boldest edges). (Adapted from~\cite{Darts})
 }
 \label{fig:intro_darts}
\end{figure}

One important application of NAS is to design hardware efficient deep models under various constraints, such as memory footprint, inference latency, and power consumption~\cite{tan2019mnasnet}. Roughly, existing NAS approaches can be categorized into three groups as shown in Figure~\ref{fig:nas_compare}: multi-shot NAS, one-shot NAS and zero-shot NAS. Multi-shot NAS methods involve training multiple candidate networks and are therefore time-consuming. It can take from a few hundred GPU hours~\cite{mao2019learning} to thousands of GPU hours~\cite{chiang2019cluster} in multi-shot NAS methods. One-shot NAS methods alleviate the computational burden by sharing candidate operations via a hyper-network~\cite{Darts,xu2019pcdarts,dong2019searchingdag,zela2019understandingrobust,chen2019progressive,cai2018proxylessnas,Cai2020Once-for-All}.
As shown in Figure~\ref{fig:intro_darts}, one-shot NAS only needs to train a single hyper-network instead of multiple candidate architectures whose number is usually exponentially large. The orders of magnitude reduction in training time enables differentiable search to achieve competitive accuracy against multi-shot NAS, but with much lower search costs~\cite{Darts}. 

\begin{figure}[tb]
 \centering
 \includegraphics[width=0.45\textwidth]{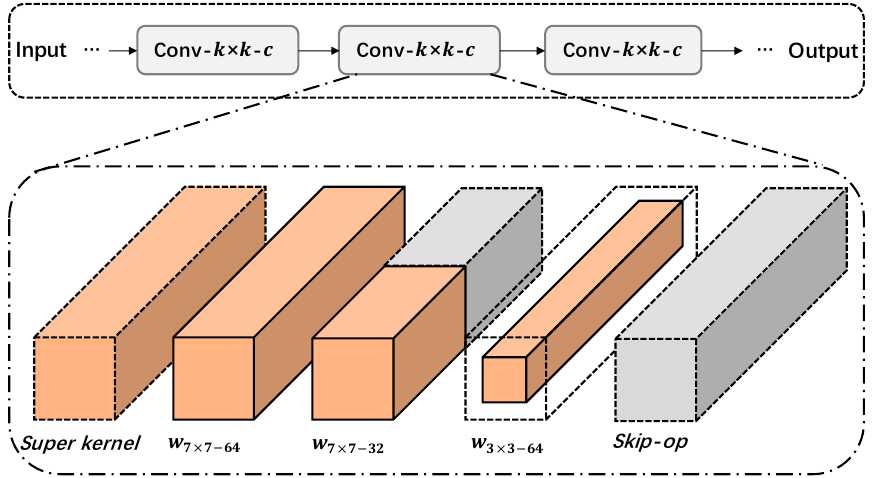}
 \caption{Illustration of weight-sharing mechanism. The parameters of relatively simple operations are obtained from complex operations, \textit{i.e.,} super kernel. As shown, different operations share the parameters from the super kernel.
 (Adapted from~\cite{single_path_nas})}
 \label{fig:intro_ws}
\end{figure}

Nevertheless, naively merging all candidate operations into a hyper-network is not efficient because the parameters of all operations need to be stored and updated during the search process. Consequently, the \textit{weight-sharing} methods improve the search efficiency of NAS even further~\cite{hyper_nas,single_path_nas,chu2021fairnas,guo2020single,chen2019fasterseg}. As shown in Figure~\ref{fig:intro_ws}, the key idea of weight-sharing NAS is to share the parameters across different operations. Next, at each training step, a sub-network is sampled from the hyper-network and then the updated parameters are copied back to the hyper-network. By sharing the parameters of various sub-networks, this differentiable search approach significantly reduces the search costs to a few or tens of GPU hours~\cite{single_path_nas}.

Though the differentiable search and weight-sharing have significantly improved the time efficiency of NAS, training is still required in one-shot NAS methods. In the last few years, the \textit{zero-shot} NAS has been proposed to liberate NAS from parameter training entirely~\cite{tfnas1,tfnas2,tfnas3nlp,tfnas4vit,tfnas5,tfnas6,tfnas7,tfnas8,tfnas12,tfnas14}.

Compared to multi-shot and one-shot methods, zero-shot NAS has the following major advantages: (\textit{i}) \textbf{Time efficiency}: zero-shot NAS utilizes some proxy as the model's test accuracy to eliminate the model training altogether during the search stage. Compared to model training, the computation costs of these proxies are much more lightweight. Therefore, zero-shot NAS can significantly reduce the costs of NAS while achieving comparable test accuracy as one-shot and multi-shot NAS approaches (see Figure~\ref{fig:nas_compare}). (\textit{ii}) {\textbf{Interpretability}}: Clearly, the quality of the accuracy proxy ultimately determines the performance of zero-shot NAS. The design of an accuracy proxy for zero-shot NAS is usually inspired by some theoretical analysis of deep neural networks thus deepening the theoretical understanding of why certain networks may work better. For example, Bhardwaj \etal developed the first zero-shot NAS approach by analyzing the topological properties of deep networks~\cite{nnmass}; some recent approaches use the number of linear regions to approximate the complexity of a deep neural network~\cite{tf_nas}. Moreover, the connection between the gradient of a network at random initialization and the accuracy of that network after training are widely explored as proxies of the model's test accuracy in zero-shot NAS~\cite{tf_nas1}.

Based on these overarching observations, this paper aims to comprehensively analyze existing hardware-aware zero-shot NAS methods. Starting from the theoretical foundations of deep learning, we first investigate various proxies of test accuracy and their theoretical underpinnings. Then, we introduce several popular benchmarks for evaluating zero-shot NAS methods. Moreover, we demonstrate their effectiveness when applied to hardware-aware NAS; notably, we reveal fundamental limitations of existing proxies. Finally, we discuss several potential research directions for hardware-aware zero-shot NAS. Overall, this paper makes the following contributions:

\begin{itemize}

 \item We review existing proxies for zero-shot NAS and provide theoretical insights behind these proxies. We categorize the existing accuracy proxies into (\textit{i}) gradient-based proxies and (\textit{ii}) gradient-free proxies. 
 \item We conduct direct comparisons of various zero-shot proxies against two naive proxies, \textit{i.e.,} \#Params and \#FLOPs, and reveal a fundamental limitation of many existing proxies: they correlate much worse with the test accuracy in constrained search settings (\textit{i.e.,} when considering only networks of high accuracies) compared to unconstrained settings (\textit{i.e.,} considering all architectures in the given search space). 
 \item We further conduct a thorough study including proxy design, benchmarks, and real hardware profiling for zero-shot NAS. We show that a few proxies have a better correlation with the test accuracy than these two naive proxies (\#Params and \#FLOPs) on the top-performing architectures such as ResNets and MobileNets.
 \item We discuss the limitations of existing zero-shot proxies and NAS benchmarks; we then outline a few possible directions for future research. 
\end{itemize}


In comparison to other existing zero-shot NAS surveys~\cite{he2021automlsurvey1,wistuba2019survey3,ren2021comprehensivesurvey4,liu2021survey5,wsnassurvey6,hwnassurvey7,colin2022adeeperlook,zeroshotnassurvery,tfnas_survet_revise}, we not only cover all existing proxies, but also provide a deep analysis of the theoretical underpinning behind them. We believe that understanding the theoretical design considerations behind these proxies is very important for future improvements.
{\color{black}Additionally, this is the first work to comprehensively compare these zero-shot proxies on large scale tasks like ImageNet-1K classification, COCO object detection, and ADE20K semantic segmentation. Furthermore, we are the first to explore the potential applicability of these zero-shot proxies to Vision Transformers.}
Last but not least, we have conducted detailed comparisons for the first time when applying zero-shot NAS in hardware-aware scenarios. This is crucial for deploying the zero-shot approaches in practice, especially for edge-AI applications.

The remaining paper is organized as follows. 
We introduce zero-shot proxies in Section~\ref{sec:proxy}. Section~\ref{sec:NASBench} surveys
existing NAS benchmarks. Hardware performance predictor is presented in Section~\ref{sec:hw_model}. We evaluate various zero-shot proxies under diverse settings in Section~\ref{sec:experiment} and point out future research directions. We conclude the paper in Section~\ref{sec:conclusion}.

\section{Zero-Shot Proxies}\label{sec:proxy}
The goal of zero-shot NAS is to design proxies that can rank the accuracy of candidate network architectures at the initialization stage, \textit{i.e.,} without training, such that we can replace the expensive training process in NAS with some computation-efficient alternatives. Hence, the proxy for the accuracy ranking is the key factor of zero-shot NAS. 

\begin{table*}
\centering
\caption{The symbols used in this paper and their corresponding meaning.}\label{tab:proxy_symbols}
\scalebox{1.0}{
\begin{tabular}{cc|cc}
\toprule
\textbf{Symbol} & \textbf{Meaning} & \textbf{Symbol} & \textbf{Meaning} \\ \midrule
 $\vx$ & Input samples & $\hat{y}$ & Ground truth (labels) \\ \midrule
 $f$ & A given deep network & $D$ & The number of layers of a given network \\ \midrule
 $f_{e}$ & A network w/o final pooling and FC layers & $y$ & The output of a given model \\ \midrule
 $\mathcal{L}$ & Loss function & $L$ & Loss values \\ \midrule 
 $\boldsymbol{\Theta}$ & All parameters of a given network & $\boldsymbol{\theta}_i$ & Parameters vector of the $i$-th layer \\ \midrule
 $\mH_i$ & Hessian matrix of the $i$-th layer & $\vz_i$ & The output vector of layer $i$ \\ \bottomrule
\end{tabular}
}
\end{table*}

\begin{table*}
\centering
\caption{\color{black} Categorization of zero-shot proxies. Based on whether or not the proxy relies on gradients, there are gradient-based and gradient-free approaches. We also categorize existing proxies by their theoretical underpinning (cf. Section \ref{sec:theoretical_underpin}). An empty cell indicates the proxy is not in that category. }\label{tab:summary_proxy}
\scalebox{0.87}{\color{black}
\begin{tabular}{c|c|c|c|c|c|c|c|c|c|c|c|c}
\toprule
\textbf{Proxy} & \textbf{Grad\_norm} & \textbf{SNIP} & \textbf{Synflow} & \textbf{GraSP} & \textbf{GradSign} & \textbf{Fisher} & \textbf{Jacob\_cov} & \textbf{NTK\_Cond} & \textbf{Zen-score} & \textbf{\#LR} & \textbf{Logdet} & \textbf{NN-Mass} \\ \toprule
\textbf{Gradient-free}  &   &   &   &   &   &  &   &   & & \textbf{\large \checkmark}  &  \textbf{\large \checkmark}  & \textbf{\large \checkmark} \\ \midrule
\textbf{Gradient-based} & \textbf{\large \checkmark} & \textbf{\large \checkmark} & \textbf{\large \checkmark} & \textbf{\large \checkmark} & \textbf{\large \checkmark} & \textbf{\large \checkmark}  & \textbf{\large \checkmark} & \textbf{\large \checkmark}  &\textbf{\large \checkmark}   &   & &   \\ \toprule
\thead{\textbf{Trainability}\\ \textbf{\&Convergence}} & \textbf{\large \checkmark} & \textbf{\large \checkmark} & \textbf{\large \checkmark} & \textbf{\large \checkmark} & \textbf{\large \checkmark}  &   & \textbf{\large \checkmark} & \textbf{\large \checkmark}  &   &   &   & \textbf{\large \checkmark} \\ \midrule
\thead{\textbf{Expressive}\\ \textbf{Capacity}}  &   &   &   &   &   & \textbf{\large \checkmark} &   &   & \textbf{\large \checkmark} & \textbf{\large \checkmark} & \textbf{\large \checkmark} & \textbf{\large \checkmark}\\ \midrule
\thead{\textbf{Generalization}\\ \textbf{Capacity}}  &   &   &   &   &   &   &   & \textbf{\large \checkmark}  &   &   &   &   \\ \bottomrule
\end{tabular}
}
\end{table*}
{\color{black}
\subsection{Theoretical Underpinning of Proxies}\label{sec:theoretical_underpin}

Before we dive deep into the details of existing zero-shot proxies, let us first establish the foundational principles for designing a good zero-shot proxy. Indeed, an ideal accuracy proxy should address three primary aspects~\cite{chen2023_understand,chen2023no}:

\begin{itemize}
\item \textbf{Expressive Capacity}:
The proxy should reflect how well the deep network can capture and model complex patterns and relationships within the data, which can be crucial for complex tasks like large-scale datasets (e.g., ImageNet-1K and COCO) ~\cite{express_power,express_survey}.
\item \textbf{Generalization Capacity}: The proxy should also reflect the network ability to generalize from the training data to unseen or out-of-distribution data. A network with a high generalization capacity should not only perform well on the training data but also on new examples, indicating that it has learned meaningful, transferable representations~\cite{general_capacity1,general_capacity2,general_capacity3}.
\item \textbf{Trainability and Convergence}:
The proxy should also indicate how quickly the network converges to a desirable performance level. Faster convergence indicates that the network is efficiently adapting to the training data and task at hand, which is essential for practical applications since training is typically expensive~\cite{train_converge1,train_converge2,train_converge3}.
\end{itemize}
In short, a good zero-shot proxy for deep network accuracy should provide insights into the network capacity to learn complex representations, generalize to unseen samples, and train to converge to minimal loss values. However, as shown in Table~\ref{tab:summary_proxy}, most existing proxies tend to target only one of these aspects. This narrow focus results in outcomes that often fail to outperform some naive proxies, like \#Params or \#FLOPs; we empirically verify this observation in Section~\ref{sec:experiment}. 
}

In this paper, we categorize the existing zero-shot proxies as follows: depending on whether or not the gradients are involved in the proxy calculation, the existing accuracy proxies fall into two major classes: (i) gradient-based accuracy proxy and (ii) gradient-free accuracy proxy (summarized in Table~\ref{tab:summary_proxy}). The symbols used in this section and their corresponding meaning are summarized in Table~\ref{tab:proxy_symbols}. 
\subsection{Gradient-based accuracy proxies}\label{sec:gradient_proxy}
We first introduce several similar proxies derived from the gradient over parameters of deep networks.

\subsubsection{Gradient norm}
The gradient norm is the sum of norms for each layer's gradient vector~\cite{tf_nas1}. 
To calculate the gradient norm, we first input a mini-batch of data into the network and then propagate the loss values backward. 
Next, we calculate the $\ell_2$-norm of each layer's gradient and then add them up for all the convolution and linear layers of the given network. Formally, the definition of gradient norm $G$ is as follows:
\begin{equation}
G\triangleq \sum_{i=1}^{D} \left\|\nabla_{\boldsymbol{\theta}_i}L\right\|_2
\end{equation}
where $D$, $\boldsymbol{\theta}_i$ and $L$ are, the number of layers, the parameter vector of the $i$-th layer of a given network and $L$ is the loss values, respectively. 

\subsubsection{SNIP}
The gradient norm only measures the property of the gradient's propagation for a given network. To jointly measure the parameter importance both in forward inference and gradient propagation, SNIP consists of multiplying the value of each parameter and its corresponding gradient~\cite{lee2018snip}. Formally, SNIP is defined as below:
\begin{equation}
\mathrm{SNIP}\triangleq\sum_{i}^{D}\left|\langle \boldsymbol{\theta}_i,\nabla_{\boldsymbol{\theta}_i}L \rangle\right|
\end{equation}
where $\langle\cdot,\cdot \rangle$ represents the inner product; $D$, $\boldsymbol{\theta}_i$ and $L$ are, the number of layers, the parameter vector of the $i$-th layer of a given network and $L$ is the loss values, respectively. 

\subsubsection{Synflow}
Similar to SNIP, Synflow consists of maintaining the sign of the SNIP proxy~\cite{synflow}:
\begin{equation}
\mathrm{Synflow}\triangleq\sum_{i}^{D}\langle \boldsymbol{\theta}_i,\nabla_{\boldsymbol{\theta}_i}L \rangle\
\end{equation}

\subsubsection{GraSP}
The three proxies mentioned above only take the first-order derivatives of neural networks into account. The GraSP proxy considers both the first-order and second-order derivatives of neural networks~\cite{grasp}. Specifically, GraSP is defined by the inner product of the parameters and the product of the Hessian matrix and the gradients:
\begin{equation}
\sum_{i}^{D}-\langle\mH_i\nabla_{\boldsymbol{\theta}_i}L,\boldsymbol{\theta}_i\rangle
\end{equation}
where $\mH_i$ is Hessian matrix of the $i$-th layer.

There are multiple theoretical analyses for the above three proxies. Specifically, Synflow and SNIP have been proven to be layer-wise constants in linear networks during the back-propagation process~\cite{lee2018snip,synflow}. Moreover, several works show that Synflow and GraSP are different approximations of the first-order Taylor expansions of deep neural networks~\cite{grasp,nvidiapruning}. We remark that Taylor expansions of a deep network can identify the parameters that contribute the most to the loss values; thus, it can measure the importance of parameters.

{\color{black}
\subsubsection{GradSign}

Given an input batch with $B$ input samples $\{\vx_1, \vx_2, ..., \vx_B\}$, GradSign is defined as follows~\cite{gradsign}:
\begin{equation}
 \mathrm{GradSign}\triangleq \sum_{\theta_k\in\bm{\Theta}}\bigg|\sum_{i=1}^B\text{sign}[\nabla_{\theta_k}\mathcal{L}\left(f(\vx_i),y_i\right)]\bigg|
\end{equation}
Essentially, GradSign assesses the uniformity across multiple training samples for each parameter, and then adds them up as the final proxy value.
It has been proven that GradSign serves as an approximation of the training loss following the training phase~\cite{gradsign}. More specifically, a higher value of GradSign is indicative of a diminished training loss. Consequently, GradSign measures the convergence properties inherent in deep neural networks.

}

Besides the gradient over parameters, the gradient over each layer's activation is also explored to build the accuracy proxy as shown below. 
\subsubsection{Fisher information}

Fisher information of a neural network can be approximated by the square of the activation value and their gradients~\cite{fisher_prune_first,fisher}: 
\begin{equation}
\sum_i^{D}\langle \nabla_{\vz_i}L, \vz_i\rangle^2
\end{equation}
where $\vz_i$ is the feature map vector of the $i$-th layer of a given network. 

Previous works show that a second-order approximation of Taylor expansion in a neural network is equivalent to an empirical estimate of the Fisher information~\cite{fisher}. Hence, measuring the Fisher information of each neuron/channel of a given network can reflect the importance of these neurons/channels.

\subsubsection{Jacobian covariant}
Besides the gradient over parameters and activations, the Jacobian covariant (Jacob\_cov) leverages the gradient over the input data $\vx$~\cite{tfnas15jacob,tfnas11logdet}. To calculate the Jacob\_cov proxy, given an input batch with $B$ input samples $\{\vx_1, \vx_2, ..., \vx_B\}$, the gradients matrix $\mJ$ of the output results $\{y_1, y_2, ..., y_B\}$ w.r.t. these inputs are first computed:
\begin{equation}
 \mJ=(\nabla_{\vx_1}y_1, \nabla_{\vx_2}y_2, ..., \nabla_{\vx_B}y_B)^T
\end{equation}
Next, the raw covariance matrix is generated as:
\begin{equation}
 \mG = (\mJ - \mM )(\mJ- \mM )^T
\end{equation}
where $M_{i,j} =\frac{1}{B}\sum_{n=1}^{B}J_{i,n}$. Then the raw covariance matrix is normalized to get the real covariance matrix $\bm{\Gamma}$:
\begin{equation}
 \bm{\Gamma}_{ {i,j} } = \frac{ G_{i,j} }{ \sqrt{ G_{i,i} G_{j,j}} }
\end{equation}
where $\Gamma_{ {i,j} }$ denotes the entries of $\bm{\Gamma}$.
\noindent Let $\lambda_1\leq \lambda_2\leq... \leq\lambda_B$ be the $B$ eigenvalues of $\bm{\Gamma}$; then the Jacobian covariant is generated as follows: 
\begin{equation}
\mathrm{Jacob\_cov}\triangleq -\sum_{i=1}^{B}\left[\left(\lambda_i+\epsilon\right) + \left(\lambda_i+\epsilon\right)^{-1}\right]
\end{equation}
where $\epsilon$ is a small value used for numerical stability.
As discussed in~\cite{tfnas15jacob,tfnas11logdet}, Jacob\_cov can reflect the expressivity of deep networks thus higher Jacob\_cov values indicate better accuracy.

\subsubsection{Zen-score}
Zen-score is a new proxy for a given model~\cite{tfnas10zen,tfnas13zendet}. The Zen-score is defined as:
\begin{equation}\label{eq:zenscore}
 \begin{aligned}
\log\mathbb{E}_{\vx,\bm{\epsilon}}&\left(\left\|f_e(\vn)- f_e(\vn + \alpha\bm{\epsilon})\right\|_F\right) +\sum_{k,i} \log\left(\sqrt{\frac{\sum_j\sigma_{ij}^k}{Ch_i}}\right),\\
 &\vx\sim\mathcal{N}(0,\mI)
\end{aligned}
\end{equation}
where, $\vn$ is a sampled Gaussian random vector, $\bm{\epsilon}$ is a small input perturbation, $\left\|\cdot\right\|_F$ indicates the Frobenius norm, $\alpha$ is a tunable hyper-parameter, $Ch_i$ is the number of channels of the $i$-th convolution layer, and $\sigma_{ij}^k$ is the variance of the $i$-th layer's $j$-th channels for the $k$-th samples in an input batch data. As shown in Eq.\ref{eq:zenscore}, Zen-score measures model expressivity by averaging the Gaussian complexity under randomly sampled $x$ and $\epsilon$. We note that this is equivalent to computing the expected gradient norm of $f$ with respect to input $x$ instead of network parameters. Hence, Zen-score measures the expressivity of neural networks instead of their trainability: networks with a higher Zen-score have a better expressivity and thus tend to have a better accuracy.

\subsubsection{NTK Condition Number}
Neural Tangent Kernel is proposed to study the training dynamics of neural networks~\cite{ntkothers}. More precisely, given two input samples $\vx_1$ and $\vx_2$, NTK is defined as:
\begin{equation}
 \kappa\left(\vx_1,\vx_2\right)=\mJ(\vx_1)\mJ(\vx_2)
\end{equation}
where $\mJ(\vx)$ is the Jacobian matrix evaluated at the sample $\vx$~\cite{ntkraw}. Lee \etal 
prove that the training dynamics of wide neural networks can be solved as follows~\cite{ntkode}:
{\color{black}
\begin{equation}\label{eq:ntk_raw}
 \mu_t(\mX_{})=\left(\mI-e^{-\eta t\mathcal{\mK}\left(\mX, \mX\right)}\right)\vy
\end{equation}
where $t$ denotes the training step; $\mu_t$ represents the output expectations at training step $t$; $\mX\in \mathbb{R}^{m\times d}$ and $\vy\in \mathbb{R}^{m}$ are the training input having $m$ samples with $d$ dimensions per sample, and their corresponding labels, respectively; $\eta$ is the learning rate. $\mathcal{\mK}\left(\mX, \mX\right)\in \mathbb{R}^{m\times m}$ is the NTK for these input data. 
By conducting the eigendecomposition of Eq.~\ref{eq:ntk_raw}, the $i$-th dimension in the eigenspace of output expectation can be written as follows:
\begin{equation}\label{eq:ntk_eigen}
 \mu_t(\mX_{i})=\left({\mI}-e^{-\eta\lambda_i t}\right)\vy_{i}, i=\{1,2,...,m\}
\end{equation}
where $\lambda_1\leq \lambda_2\leq... \leq\lambda_m$ are the eigenvalues of the NTK $\mathcal{\mK}\left(\mX_{}, \mX_{}\right)$. 
}

Therefore, a smaller difference between $\lambda_1$ and $\lambda_m$ indicates (on average) a more ``balanced" convergence among different dimensions in the eigenspace. To quantify the above observation, the NTK Condition Number (NTK\_Cond) is defined as follows~\cite{tf_nas}: 
\begin{equation}
 \mathrm{NTK\_Cond} \triangleq \mathbb{E}_{\mX_{},\bm{\Theta}}\frac{\lambda_m}{\lambda_1}
\end{equation}
 where $\bm{\Theta}$ is the randomly initialized network parameters. Chen \etal demonstrate that the NTK\_Cond is negatively correlated with the architecture’s test accuracy~\cite{tf_nas}. Hence, the networks with lower NTK\_Cond values tend to have a higher test accuracy. Similar insights are reported and leveraged in \cite{asvit} for NAS of vision transformers (ViTs).

\subsection{Gradient-free accuracy proxy}\label{sec:nograd_proxy}
Though the gradient-based proxies do not require the training process on the entire dataset, backward propagation is still necessary to compute the gradient. To entirely remove the gradient computation from the neural architecture search, several gradient-free proxies have been proposed lately. 

\subsubsection{Number of linear regions}
\begin{figure}[htb]
 \centering
 \includegraphics[width=0.45\textwidth]{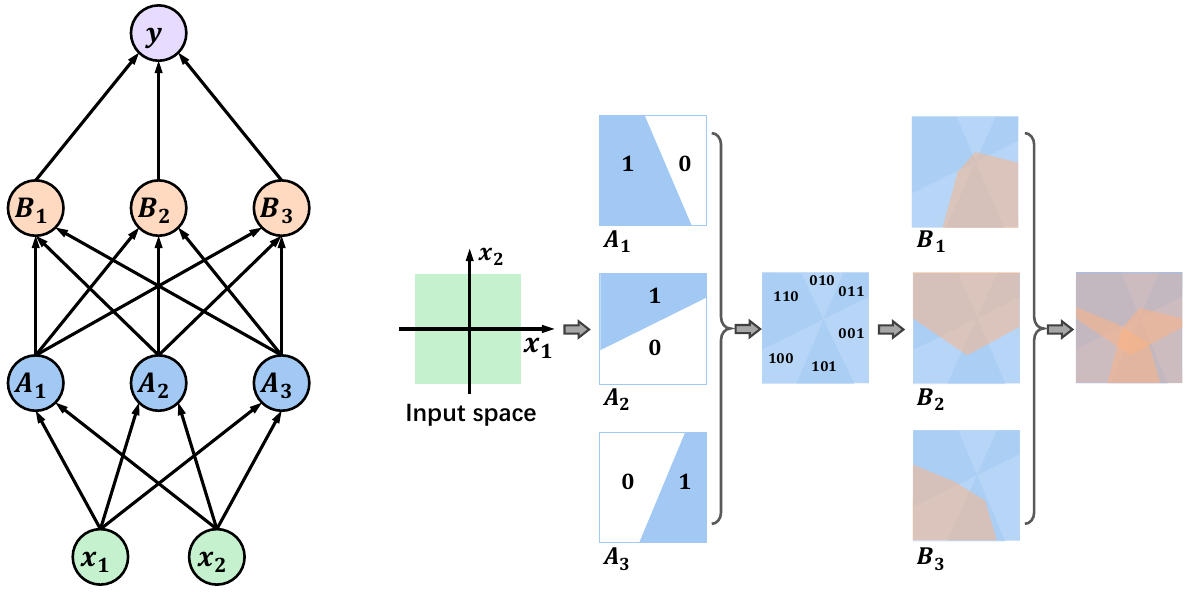}
 \caption{The illustration of Logdet proxy; $A_i, B_i, i=\{1,2,3\}$ are the neurons of a multi-layer perceptron. First, the input space is divided into several linear regions. Next, each region is encoded by a binary code; then Eq.~\ref{eq:Logdet} is applied to compute the Logdet proxy. (Adapted from~\cite{tfnas11logdet})}
 \label{fig:logdet}
\end{figure}

{\color{black} The number of linear regions in a neural network indicates the distinct sections into which the network can partition its input space}; thus, it describes the expressivity of a given network~\cite{linearregions1,linearregions2,linearregions,linearregions3}. For instance, a single-neuron perceptron with a ReLU activation function can divide its input space into two regions. 
Previous work shows that one can estimate the number of linear regions with the help of the activation patterns in the output activation matrix $\mR$~\cite{linearregions}:
\begin{equation}
 \mR=\bm{1}\cdot\bm{1}^T- \mathrm{sign}[\vz_i(\bm{1}-\vz_i)^T + (\bm{1}-\vz_i)\vz_i^T]
\end{equation}
where $\bm{1}$ is an all-one vector. Next, by removing the repeating patterns and assigning the weights to each pattern, the number of linear regions $\rho$ is as follows:
\begin{equation}
 \rho\triangleq\sum_j\frac{1}{\sum_{k}R_{j,k} }
\end{equation}
where $R_{j,k}$ is the entry of $R$. Therefore, the number of linear regions measures how many unique regions the network can divide the entire activation space into (see Figure~\ref{fig:logdet}).

\subsubsection{Logdet}
Logdet is another proxy proposed based on the number of linear regions~\cite{tfnas11logdet}:
\begin{equation}\label{eq:Logdet}
\begin{aligned}
 \mH&=\begin{bmatrix}
 N_{LR}-d_H(\vc_1,\vc_1)& \cdots & N_{LR}-d_H(\vc_1,\vc_N)\\
 \vdots & &\vdots \\
 N_{LR}-d_H(\vc_N,\vc_1)& \cdots & N_{LR}-d_H(\vc_N,\vc_N) \\
\end{bmatrix}\\
 &\ \ \ \ \ \ \ \ \ \ \ \ \ \ \ \ \ \ \ \ \ \ \ \ \mathrm{Logdet}\triangleq \log|\mH|
\end{aligned}
\end{equation}
where $N_{LR}$ is the total number of linear regions, $d_H$ is the Hamming distance, and $\vc_i$ is the binary coding vector of the $i$-th linear region as shown in Figure~\ref{fig:logdet}. Previous work shows that networks with a higher Logdet at initialization tend to have higher test accuracy after training~\cite{tfnas11logdet}.


\subsubsection{Topology inspired proxies}\label{sec:nnmass}
The very first pioneering work behind theoretically-grounded, training-free architecture design was done by Bhardwaj et al.~\cite{nnmass}. While the above proxies are proposed for a general search space, \textit{i.e.,} without any constraints on the candidate architectures, as discussed later, these general-purpose proxies are not better than some naive proxies, \textit{e.g.,} the number of parameters (\#Params) of a model. To design better accuracy proxies than \#Params, Bhardwaj et al.~\cite{nnmass} constrained the search space to specific topologies, e.g., DenseNets, ResNets, MobileNets, etc., and theoretically studied how network topology influences gradient propagation. Inspired by the network science, NN-Mass is defined as follows~\cite{nnmass}:
\begin{equation}
\begin{aligned}
 \rho_{c}\triangleq&\frac{\mathrm{\#Actual\ skip\ connections\ of\ cell}\ c}{\mathrm{\#Total\ possible\ skip\ connections\ of\ cell}\ c}\\
 &\quad \quad\quad \text{NN-Mass}\triangleq\sum_{\mathrm{each\ cell}\ c}\rho_c w_c d_c 
\end{aligned}
\end{equation}
where $w_c$ and $d_c$ are the width and depth values of a cell\footnote{A cell represents a group of layers with the same width values or commonly used blocks in CNN, \textit{e.g.,} Basic/Bottleneck blocks in ResNet, and Inverted bottleneck blocks in MobileNet-v2.}, respectively. Bhardwaj~\etal prove that higher NN-Mass values indicate better trainability of networks and faster convergence rate during training~\cite{linearregions1}. Moreover, they also show that networks with higher NN-Mass values tend to achieve a higher accuracy. NN-Mass has also been used to perform training-free model scaling to significantly improve accuracy-MACs tradeoffs compared to highly accurate models like ConvNexts~\cite{bhardwaj2022restructurable}. In~\cite{bhardwaj2022restructurable}, Bhardwaj \etal show the connection between NN-Mass and expressive power of deep networks for ResNet-type networks.

As an extension of NN-Mass, NN-Degree is proposed by relaxing the constraints on the width of networks. Formally, NN-Degree is defined as follows~\cite{flash}:
\begin{equation}
 \text{NN-Degree}=\sum_{\mathrm{each\ cell}\ c}(w_c + \frac{\mathrm{\#Actual\ skip\ connections}}{\mathrm{\#Total\ input \ channels}})
\end{equation}
where $w_c$ is the average width value of a cell $c$. Similarly to NN-Mass, NN-Degree has shown a high positive correlation with the test accuracy. 

Lately, Chen \etal developed another principled approach for understanding of a neural network connectivity patterns based on its capacity or trainability~\cite{chen2022deep}. Specifically, they theoretically characterized the impact of connectivity patterns on the convergence of deep networks under gradient descent training with fine granularity, by assuming a wide network and analyzing its Neural Network Gaussian Process (NNGP) \cite{lee2018deep}. Chen \etal also prove that how the spectrum of an NNGP kernel propagates through a particular connectivity pattern would affect the bounds of the convergence rates. On the practical side, they show that such NNGP-based characterization could act as a simple filtration of ``unpromising" connectivity patterns, to significantly accelerate the large-scale neural architecture search without any overhead.

{\color{black}
\subsection{Summary}
As shown in Table~\ref{tab:summary_proxy}, most of the existing zero-shot proxies are gradient-based. We note that to calculate the gradient typically involves the backward propagation. Hence, gradient-based proxies are less efficient than gradient-free proxies. Besides, most of the gradient-based proxies (except for Fisher and Logdet), are designed to measure the trainability of deep networks. In contrast, most of the gradient-free proxies (except for NN-Mass) are indicatives of the expressive capacity of neural networks. 
Moreover, apart from NTK\_Cond, current proxies fail to quantify the generalization capacity of deep networks. Future proxy designs should address and rectify this limitation.

More importantly, as highlighted earlier, the majority of existing zero-shot proxies (with the exceptions of NTK\_Cond and NN-Mass) concentrate solely on one of three dimensions: \{expressive capacity, generalization capacity, trainability\}. This is a fundamental shortcoming, as a good neural network seamlessly integrates all three facets. We provide empirical evidence of this concern in Section~\ref{sec:experiment}.
}
\section{Benchmarks and Profiling Models}\label{sec:NASBench}
NAS benchmarks have been proposed to provide a standard test kit for fair evaluation and comparisons of various NAS approaches~\cite{siems2020NASBench301,mehta2022benchsuite,mehrotra2020benchasr,klyuchnikov2022benchnlp}. A NAS benchmark defines a set of candidate architectures and their test accuracy or hardware costs. We classify the existing NAS benchmarks as standard NAS (\textit{i.e.,} without hardware costs) and hardware-aware NAS benchmarks. Next, we introduce these two types of NAS benchmarks. 

\subsection{NAS Benchmarks}\label{sec:std_nas_bench}
\begin{figure*}
 \centering
 \includegraphics[width=0.96\textwidth]{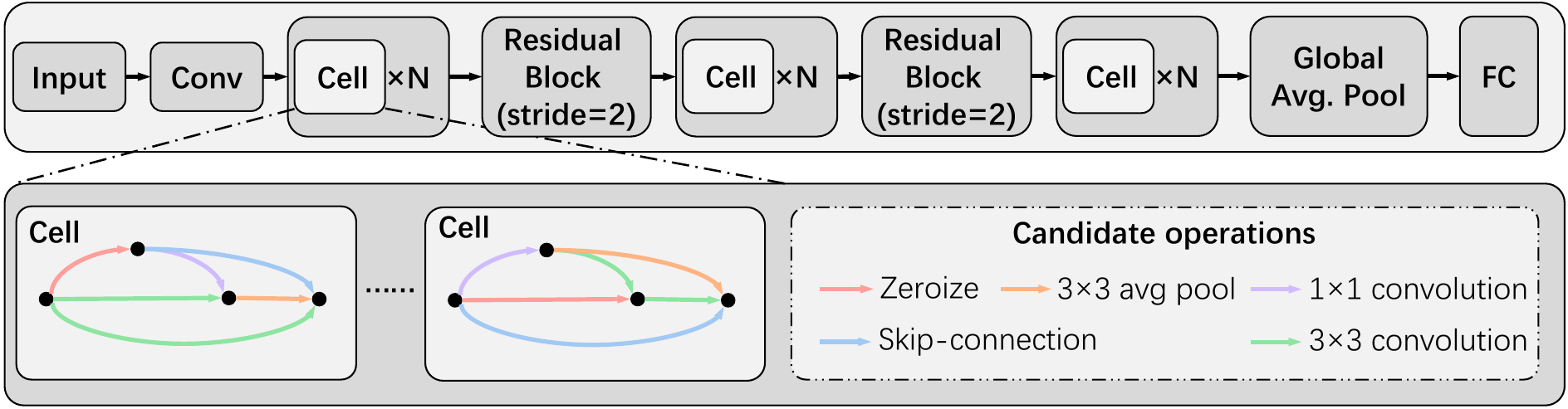}
 \caption{Search space of NASBench-201. Each architecture in the search space is built by stacking a cell multiple times; each cell can have six operations (edges in the figure) and each operation has 5 potential different options (drawn with different colors). NASBench-101 has a very similar search space with more candidate operations. (Adapted from~\cite{Darts})}
 \label{fig:nb201}
\end{figure*}
{\color{black}
We evaluate the zero-shot proxies on the following standard NAS benchmarks: 
\textbf{NASBench-101} provides 423k neural architectures and their test accuracy on the CIFAR10 dataset, where each architecture is built by stacking a cell for multiple times~\cite{ying2019NASBench101}. \textbf{NATS-Bench} contains two sub-search spaces: \textit{(i) NATS-Bench-TSS}, also known as NASBench-201; each network in NASBench-201 is also built by repeating a cell multiple times on three datasets, namely, CIFAR10, CIFAR100, and ImageNet16-120~\cite{dong2020NASBench201} (see Figure~\ref{fig:nb201} for more details); \textit{(ii) NATS-Bench-SSS} contains 32768 architectures with different width values for each layer~\cite{dong2021natsbench}\footnote{In the rest of the paper, we use the NATS-Bench to represents NATS-Bench-SSS for short.}. 
\textbf{TransNAS-Bench-101} is a benchmark dataset containing network performance on seven diverse vision tasks, including image classification, image reconstruction, and pixel-level prediction~\cite{transNAS_Bench_101} with two different sub-search spaces: (\textit{i}) A cell-level search space consisting of 4,096 unique networks with different cells; (\textit{ii}) A macro-level search space containing 3256 unique networks with different depth values.
}

\noindent\textbf{Hardware-aware NAS benchmarks.} Recent hardware-aware NAS approaches aim to jointly optimize the test performance and hardware efficiency of neural architectures. Hence, hardware-aware NAS benchmarks have been proposed by incorporating the hardware costs of networks into the search process.
HW-NAS-Bench covers the search space from both the NASBench-201 and FBNet~\cite{li2021hwNASBench}. It provides all the architectures in these two search spaces measured/estimated hardware cost (\textit{i.e.,} latency and energy consumption) on multiple types of devices. 
{\color{black}Similarly, Eagle, also known as BRP-NAS, provides a benchmark that contains latency and energy for NAS-Bench-201 networks running on up to 13 devices spanning a wide spectrum from the cloud server to the edge devices; this ameliorates the need for researchers to have access to these devices~\cite{brp_nas}. Moreover, Eagle also proposes an efficient performance estimator for measuring and predicting the performance of neural networks (cf. Section \ref{sec:hw_model}). }

\subsection{Hardware Performance Models}\label{sec:hw_model}
To incorporate the hardware-awareness into NAS, we also need to construct models to efficiently and accurately estimate the hardware performance (\textit{e.g.,} latency) of given networks. In this section, we consider latency to characterize the hardware performance and use NASBench-201 as an example to compare several representative approaches for hardware performance models.

\begin{table}[htb]
 \centering
 \caption{Comparison of representative hardware performance models. Granularity refers to the level of input features for the hardware performance models, and transferability denotes the efficiency with which the model for one hardware platform can be transferred to another. The latency is measured on Snapdragon-888's GPU with NASBench-201 on CIFAR100 dataset. }
 \label{tab:HW_MODEL}
 \scalebox{0.82}{
 \begin{tabular}{c|c|c|c|c}
 \toprule
 \textbf{Approach} & \textbf{Method} & \textbf{Granularity} & \textbf{Transferability} &\textbf{RMSE(ms)}\\\midrule
 \textbf{BRP-NAS}\cite{brp_nas} & GCN or MLP & Layer & Low & 4.6\\ \midrule
 \textbf{HELP}\cite{help_hw} & GCN or MLP& Layer & High & 0.12\\ \midrule
 \textbf{NN-Meter}\cite{nn_meter} & GCN & Kernel & Low & 1.2\\ \bottomrule
 \end{tabular}}
\end{table}

BRP-NAS is a pioneering approach that uses deep learning to build hardware performance models~\cite{brp_nas}. Specifically, BRP-NAS first converts a neural network into a directed acyclic graph by modeling each layer as an edge in a graph and modeling the input/output as nodes in the graph. Next, by using different values to present different types of layers, BRP-NAS uses a Graph Convolution Network (GCN) to build the hardware performance models. Then the model is trained with multiple networks and their real hardware performance data on the target hardware. In particular, for the networks with fixed depth, BRP-NAS can also use MLP to build the performance model. 
Though BRP-NAS can achieve good prediction results with enough training samples, there is a limitation for BRP-NAS: the performance model is trained for a specific hardware platform; if new hardware comes, one needs to repeat the entire process. 

To address the above problem, HELP builds the hardware performance models by taking the hardware information as extra input features (\textit{e.g.,} type of the hardware, number of computing elements, and the size of on-chip memory)~\cite{help_hw}. Next, HELP is trained with the latency data collected from multiple platforms, such as desktop CPU/GPU and mobile CPU/GPU. This way, if new hardware comes in, HELP only needs a few samples to conduct the fine-tuning process (typically around 10). Hence, HELP is very efficient in terms of the transferability to new hardware.
Nevertheless, both BRP-NAS and HELP are built on the layer-level analysis, which is relatively coarse for an accurate prediction.

To further improve the accuracy of performance models, NN-Meter is proposed by analyzing the neural network at a finer granularity during run-time. Specifically, NN-Meter computes the kernels of each neural network, which are originally generated during the compilation process~\cite{nn_meter}. To remove the necessity of the compilation process, NN-Meter utilizes the algorithm to automatically predict the generated kernels. Hence, as shown in Table~\ref{tab:HW_MODEL}, NN-Meter has a much higher prediction quality than both HELP and BRP-NAS.

\section{Experimental results}\label{sec:experiment}
In this section, we compare the existing proxies on multiple NAS benchmarks under various scenarios. Besides the proxies mentioned above, we also evaluate two naive proxies, \textit{i.e.,} \#Params and \#FLOPs. 

{\color{black}
\noindent\textbf{Evaluation Metrics.} We use two commonly used criteria to evaluate the correlations between different zero-shot proxies and their test accuracies across different benchmarks:
\begin{itemize}
 \item Spearman’s $\rho$. Spearman’s $\rho$ quantifies the monotonic relationships between two variables within the range of [-1, 1], where $\rho = 1$ indicates a perfect positive correlation between these two variables, while $\rho = -1$ indicates a perfect negative correlation. We use ``SPR" for short to represent Spearman’s $\rho$ in the tables and figures of this paper. 
 \item Kendall’s $\tau$. Similar to Spearman’s $\rho$, Kendall’s $\tau$ value is also within [-1, 1]. Typically, Kendall’s $\tau$ is more robust to error and discrepancies than Spearman’s $\rho$. We use ``KT" for short to represent Kendall’s $\tau$ in the tables and figures of this paper. 
\end{itemize}
In NAS, the architectures with good performance are more important than those with poor performance. Hence, we also calculate Spearman’s $\rho$ and Kendall’s $\tau$ for the architectures with test accuracy ranking top 5\% in the entire search space, which are denoted as ``SPR@Top 5\%'' and ``KT@Top 5\%'', respectively. Similarly, if we calculate Spearman’s $\rho$ and Kendall’s $\tau$ for all architectures in the search space, they are denoted as ``SPR@All'' and ``KT@All'', respectively.
}
\begin{figure}
 \centering
 \includegraphics[width=0.45\textwidth]{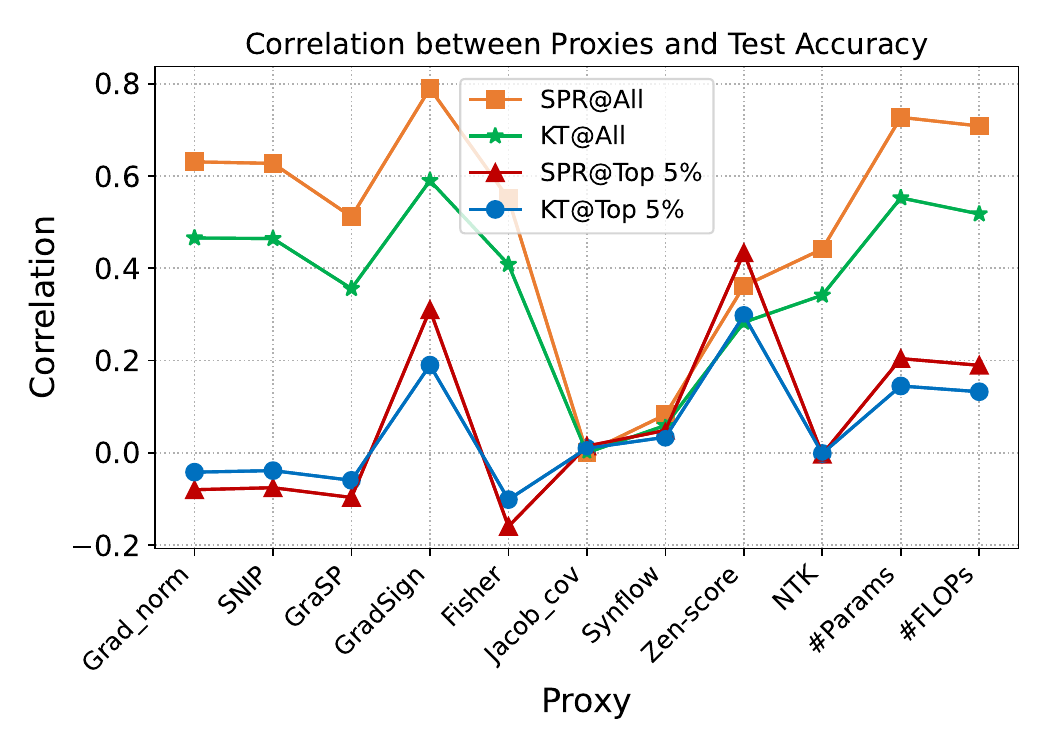}\vspace{0mm}
 \caption{The correlation between various proxies and the test accuracy on NASBench-201 search space for CIFAR-100 dataset (averaged over 5 seeds). All: all the networks in the benchmark; Top 5\%: the architectures with test accuracy ranking top 5\% in the entire search space. KT and SPR are short for Kendall's~$\tau$ and Spearman's~$\rho$, respectively (same for other figures).}
 \label{fig:proxy_201_c100}
\end{figure}

\begin{figure}
 \centering
 \includegraphics[width=0.45\textwidth]{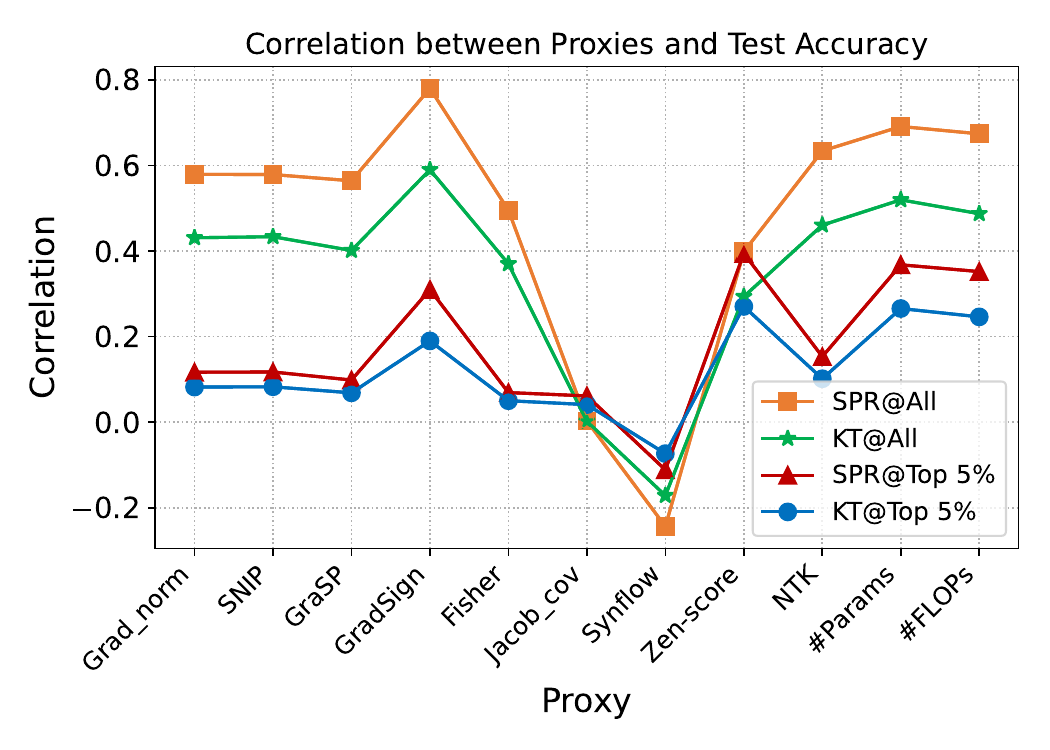}\vspace{0mm}
 \caption{The correlation between various proxies and the test accuracy on NASBench-201 search space for ImageNet16-120 dataset (averaged over 5 seeds). }
 \label{fig:proxy_201_tinyimg}
\end{figure}

\begin{figure}
 \centering
 \includegraphics[width=0.45\textwidth]{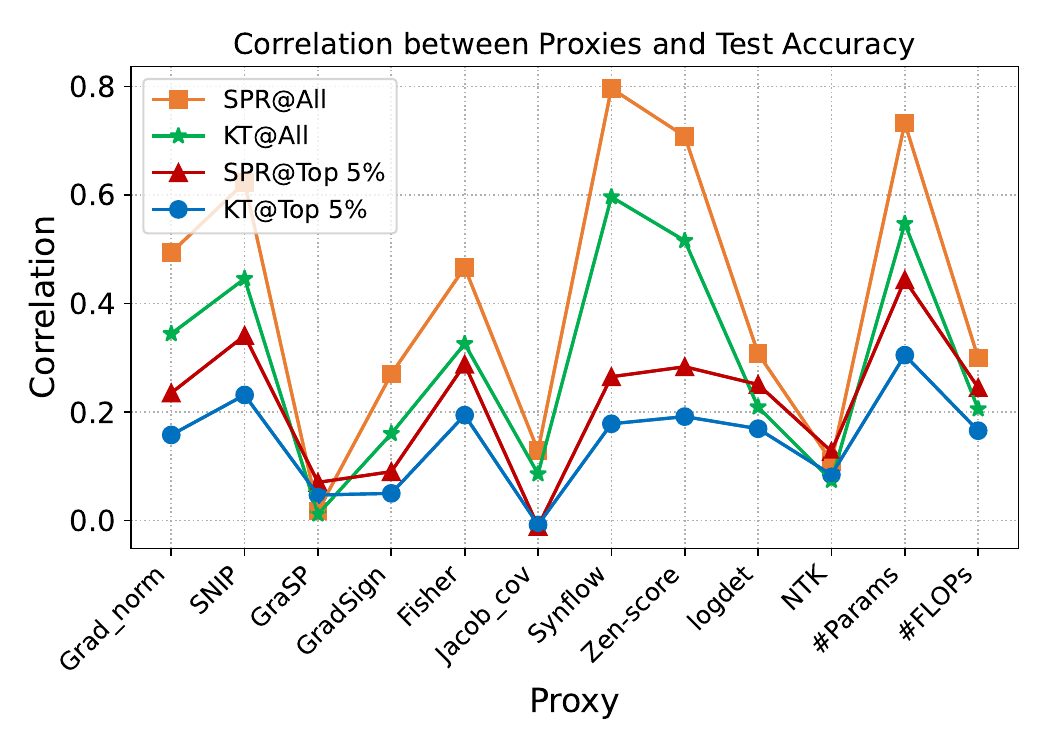} \vspace{0mm}
 \caption{The correlation between various proxies and the test accuracy on NATS-Bench search space for CIFAR100 dataset (averaged over 5 seeds). }
 \label{fig:proxy_nats_c100}
\end{figure}

\begin{figure}
 \centering
 \includegraphics[width=0.45\textwidth]{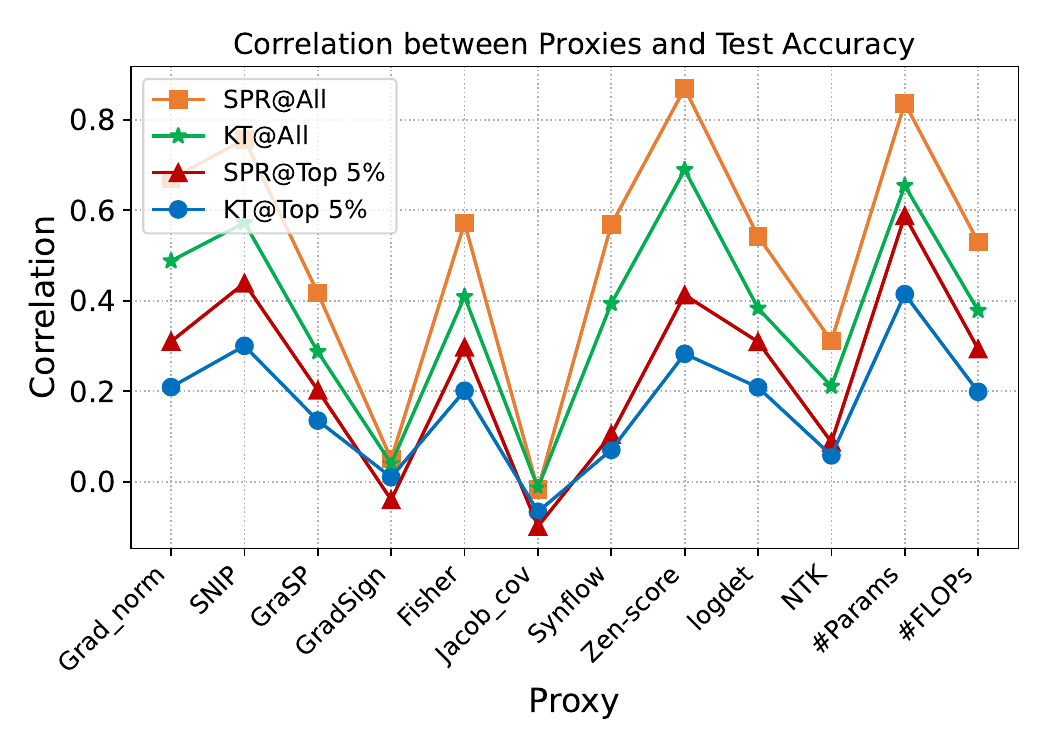} \vspace{0mm}
 \caption{The correlation between various proxies and the test accuracy on NATS-Bench search space for ImageNet16-120 dataset (averaged over 5 seeds). }
 \label{fig:proxy_nats_tinyimg}
\end{figure}

\begin{figure}
 \centering
 \begin{subfigure}[b]{0.45\textwidth}
 \centering
 \includegraphics[width=\textwidth]{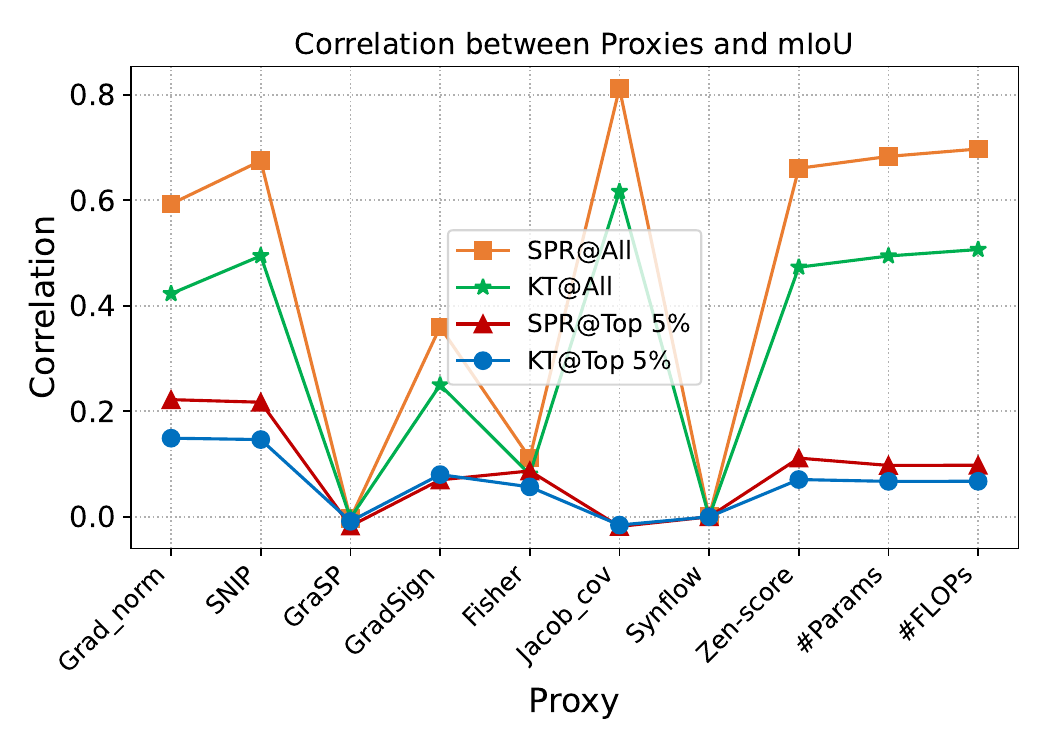}\vspace{-3.5mm}
 \caption{\color{black} Semantic Segmentation}\vspace{2mm}
 \label{fig:tnb101_semseg}
 \end{subfigure}
 
 \begin{subfigure}[b]{0.45\textwidth}
 \centering
 \includegraphics[width=\textwidth]{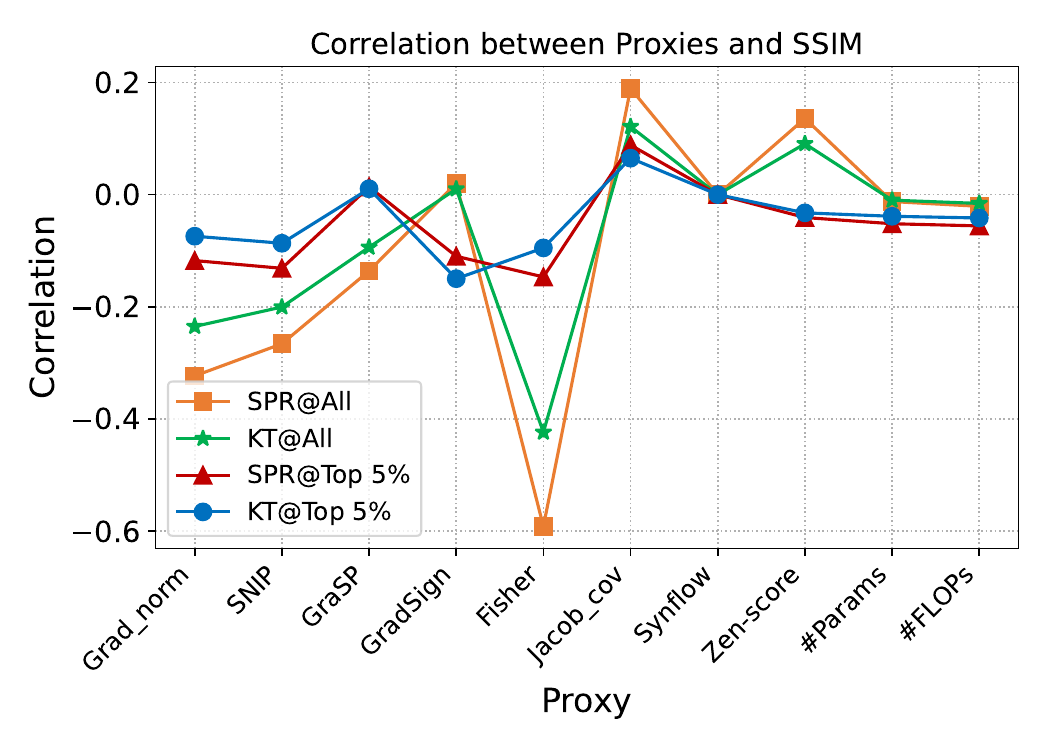}\vspace{-3.5mm}
 \caption{\color{black} Autoencoding}\vspace{2mm}
 \label{fig:tnb101_autoencoder}
 \end{subfigure}
 
 \begin{subfigure}[b]{0.45\textwidth}
 \centering
 \includegraphics[width=\textwidth]{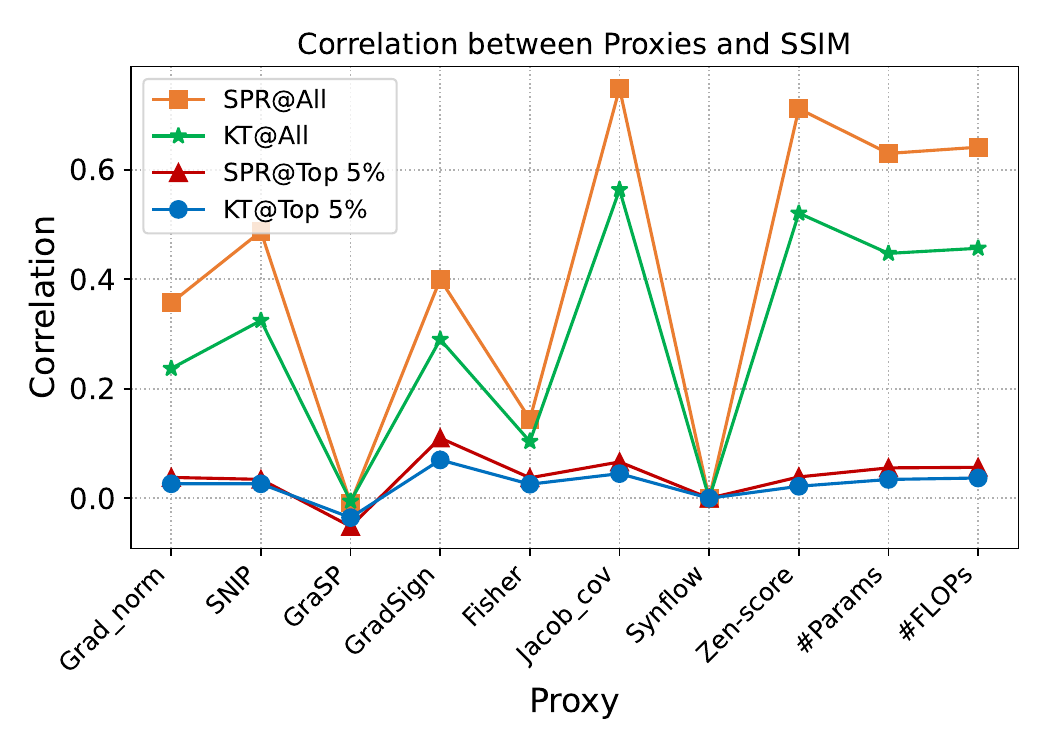}\vspace{-3.5mm}
 \caption{\color{black} Surface Normal}
 \label{fig:tnb101_normal}
 \end{subfigure}
 \caption{{\color{black}The correlation between various proxies and test performance on TransNAS-Bench-101 for Semantic Segmentation, Autoencoding, and Surface Normal tasks (averaged over 5 seeds).}}
 \label{fig:tnb101_core}
\end{figure}

\begin{figure}
 \centering
 \includegraphics[width=0.45\textwidth]{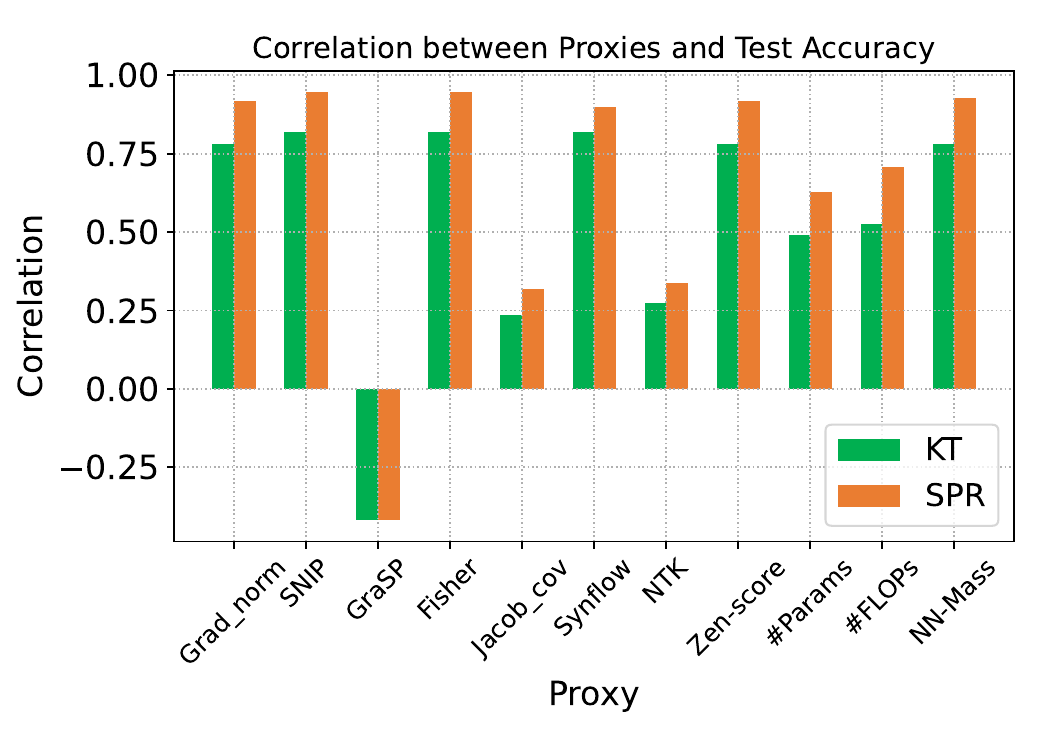}
 \caption{The correlation between various proxies and the test accuracy on a set of ResNets and Wide-ResNets for ImageNet-1K classification (averaged over 5 seeds). }
 \label{fig:nnmass_resnet}
\end{figure}

\begin{figure}
 \centering
 \includegraphics[width=0.45\textwidth]{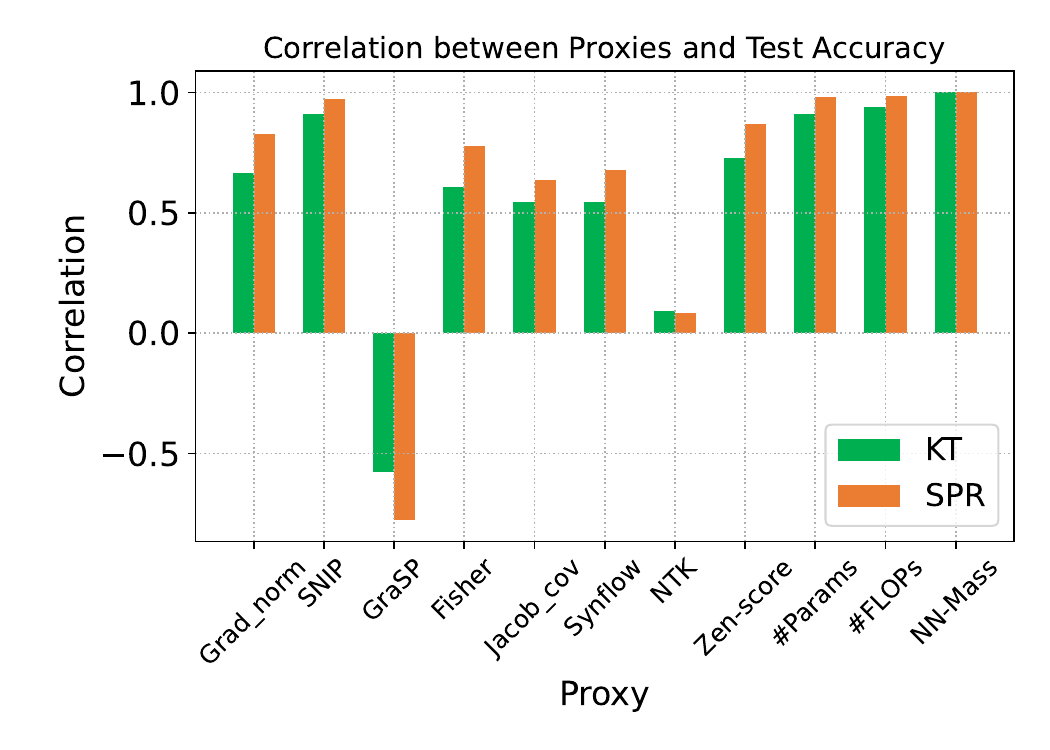}
 \caption{The correlation between various proxies and the test accuracy on a set of MobileNet-v2-based networks for ImageNet-1K classification (averaged over 5 seeds).}
 \label{fig:nnmass_mbnv2}
\end{figure}

\subsection{NAS without hardware-awareness}\label{sec:comparison_purenas}
To compare the performance of these proposed accuracy proxies, we calculate the correlation of these proxy values and the real test accuracy. We next discuss the results on two NAS benchmarks: NASBench-201 and NATS-Bench. 

\begin{table*}[]
 \centering
 \caption{The test accuracy (\%) of optimal architectures obtained by various zero-shot proxies (averaged over 5 runs) on NASBench-201 (NB201) and NATS-Bench (NB201) for CIFAR100 (C100) and ImageNet16-120 (Img16) datasets. The best results are shown with bold fonts. }
 \scalebox{0.89}{
 \begin{tabular}{ccc|c|c|c|c|c|c|c|c|c|c}
 \toprule
 & {\textbf{Proxies}} & \textbf{Ground Truth} & \textbf{Grad\_norm} & \textbf{SNIP} & \textbf{GraSP} &\textbf{GradSign} & \textbf{Fisher} & \textbf{Jacob\_cov} & \textbf{Synflow} & \textbf{Zen-score} & \textbf{\#Params} & \textbf{\#FLOPs} \\ \midrule
 \multicolumn{1}{c}{\multirow{2}{*}{\textbf{NB201}}} & \textbf{C100} & 73.51 & 60.02 & 60.02 & 60.02 & 60.02 & 60.02 & 68.89 & 62.22 & 68.10 & \textbf{71.11} & \textbf{71.11} \\ 
 \cmidrule(lr){2-13} 
 \multicolumn{1}{c}{} & \textbf{Img16} & 47.31 & 29.27 & 29.27 & 5.46 & 5.46 & 29.27 & 25.07 & 26.08 & 40.77 & \textbf{41.44} & \textbf{41.44} \\ \midrule
 \multicolumn{1}{c}{\multirow{2}{*}{\textbf{NATS}}} & \textbf{C100} & 70.92 & 48.44 & 68.36 & 57.40 & 57.40 & 53.14 & 55.04 & 66.84 & 69.92 & \textbf{70.28} & \textbf{70.28} \\ 
 \cmidrule(lr){2-13} 
 \multicolumn{1}{c}{} & \textbf{Img16} & 46.73 & 40.97 & 45.63 & 33.97 & 33.97 & 35.80 & 35.03 & 35.37 & \textbf{46.27} & 44.73 & 44.73 \\ \bottomrule
 \end{tabular}
 }
 \label{tab:optimal_acc}
\end{table*}
\subsubsection{Unconstrained search space}\label{sec:exp_unconstrained}
We first investigate the performance of zero-shot proxies for the unconstrained search spaces, \textit{i.e.,} considering all networks in the benchmarks.

\noindent\textit{NASBench-201}:
We calculate the correlation coefficients between multiple proxies and the test accuracy on CIFAR-100 and ImageNet16-120 datasets. As shown in Figure~\ref{fig:proxy_201_c100} and~\ref{fig:proxy_201_tinyimg}, the \#Params generally works best for these two datasets. Except for the \#Params, several gradient-based proxies, such as Grad\_norm, SNIP, GraSP, and Fisher, also work well.

As shown in Table~\ref{tab:optimal_acc}, we compare the neural architectures with the highest test accuracy found via various proxies. The neural architectures obtained via \#Params and \#FLOPs have the highest test accuracy on NASBench-201, which is natural and expected results given the correlation scores above.

\noindent\textit{NATS-Bench}:
Similar to NASBench-201, we calculate the correlation coefficients between these proxies and the test accuracy on CIFAR-100 and ImageNet16-120 datasets for NATS-Bench. As shown in Figure~\ref{fig:proxy_nats_c100} and~\ref{fig:proxy_nats_tinyimg}, the \#Params and Zen-score generally work best for these two datasets. 

{

\begin{table*}[t]
\centering
\caption{\color{black}The test performance of optimal architectures obtained by various zero-shot proxies (averaged over 5 runs) on TransNAS-Bench-101 benchmarks. The best results are shown with bold fonts. Here, the evaluation metric for semantic segmentation is mIoU, while the rest two use SSIM~\cite{ssim}. }
\scalebox{0.91}{\color{black}
\begin{tabular}{cc|c|c|c|c|c|c|c|c|c|c}
\toprule
\textbf{Task} & \textbf{GroundTruth} & \textbf{Gradnorm} & \textbf{SNIP} & \textbf{GraSP} & \textbf{GradSign} & \textbf{Fisher} & \textbf{Jacob\_cov}& \textbf{Synflow} & \textbf{Zen-score} & \textbf{\#Params} & \textbf{\#FLOPs} \\ \midrule
\thead{\textbf{Semantic}\\ \textbf{Segmentation}} & 94.61 & 91.66 & 94.43 & 94.53 & 90.19 & 91.89 & 94.34 & 94.46 & \textbf{94.50} & \textbf{94.50} & \textbf{94.50} \\ \midrule
\textbf{Surface Normal} & 0.59 & 0.53 & 0.53 & 0.38 & \textbf{0.57} & \textbf{0.57} & 0.55 & 0.53 & 0.55 & 0.55 & 0.55 \\ \midrule 
\textbf{Autoencoding} & 0.58 & 0.36 & 0.33 & 0.33 & 0.35 & \textbf{0.49} & 0.42 & 0.46 & 0.46 & 0.46 & 0.46 \\ \bottomrule
\end{tabular}
\label{tab:optimal_acc_tnb101}
}
\end{table*}

\color{black}
\noindent\textit{TransNAS-Bench-101}: So far, we primarily compare these zero-shot proxies on the classification tasks. To verify the effectiveness of these proxies for more diverse applications, we make comparisons for non-classification tasks selected from the TransNAS-Bench-101. We pick the largest search space TransNAS-Bench-101-Micro which contains 4096 total architectures with different cell structures. We compare these proxies under the following three non-classification tasks:
\begin{itemize}
 \item Semantic segmentation. Semantic segmentation involves classifying each pixel in an image into a predefined category or class. Unlike object detection, which identifies the bounding boxes around objects, or image classification, which assigns a single label to the entire image, semantic segmentation provides a detailed, pixel-level classification. 
 \item Surface Normal. Similar to semantic segmentation, surface normal is a pixel-level prediction task that predicts surface normal statistics.
 \item Autoencoding. Autoencoding is an end-to-end image reconstruction task that encodes an input image into a low-dimension representation vector and then reconstructs this vector into the input image.
\end{itemize}
As shown in Figure~\ref{fig:tnb101_core}, Jacob\_cov typically achieves the highest correlation for these two tasks and consistently outperforms \#Params. Besides Jacob\_cov, the Zen-score also works well and it consistently surpasses \#Params.

We also compare the neural architectures with the highest test accuracy found via various proxies. As shown in Table~\ref{tab:optimal_acc_tnb101}, the neural architectures obtained via \#Params and \#FLOPs consistently have the highest or second-highest test performance on TransNAS-Bench-101.

}

Overall, it appears that none of these proposed accuracy proxies consistently have a higher correlation with the test accuracy compared to \#Params and \#FLOPs for these two NAS benchmarks. 

\subsubsection{Constrained search space}\label{sec:top5}
We note that the architectures with high accuracy are much more important than those networks with low test accuracy. Hence, we calculate the correlation coefficient for the architectures with test accuracy ranking top 5\% in the entire search space. 
Figure~\ref{fig:proxy_201_c100} and~\ref{fig:proxy_201_tinyimg} show that, compared to ranking without constraints (\textit{i.e.,} considering all architectures), the correlation score has a significant drop except for the Zen-score on NASBench-201. Similarly, on NATS-Bench, Figure~\ref{fig:proxy_nats_c100} and~\ref{fig:proxy_nats_tinyimg} show that most of the proxies have a significant correlation score drop when constrained to the top 5\% networks in the search space, including \#Params and \#FLOPs. {\color{black} By switching to non-classification tasks, we observe a similar trend in Figure~\ref{fig:tnb101_core}, i.e., there's a significant correlation score drop under these constrained scenarios. }

This drop in correlation score for the top 5\% of networks means the zero-shot NAS is more likely to miss the optimal or near-optimal networks. Table~\ref{tab:optimal_acc} shows that there is a big accuracy gap between the ground truth and the networks obtained by each proxy. results become even worse with a search that has more relaxed hardware constraints (see Sec~\ref{sec:comparison_hwnas}). 

\begin{figure}
 \centering
 \includegraphics[width=0.45\textwidth]{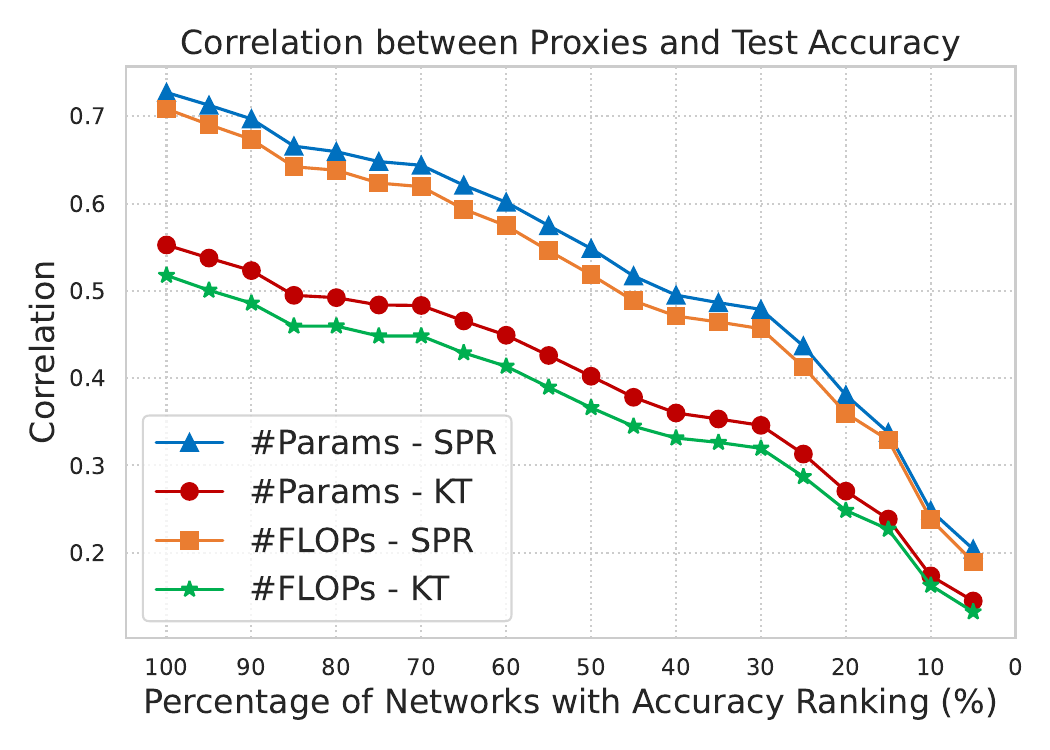}
 \caption{The correlation between \#Params \& \#FLOPs and the test accuracy under various ratios of networks on NASBench-201 for CIFAR100 dataset (averaged over 5 seeds). 20\% means computing the correlation scores only for the networks whose test accuracy ranks top 20\% in the benchmark; 100\% means considering all the networks in the benchmark (same for Figure~\ref{fig:ratio_nats_img}). From left to right, the search space is more and more constrained to neural architectures with high accuracy. }
 \label{fig:ratio_201_c100}
\end{figure}

\begin{figure}
 \centering
 \includegraphics[width=0.45\textwidth]{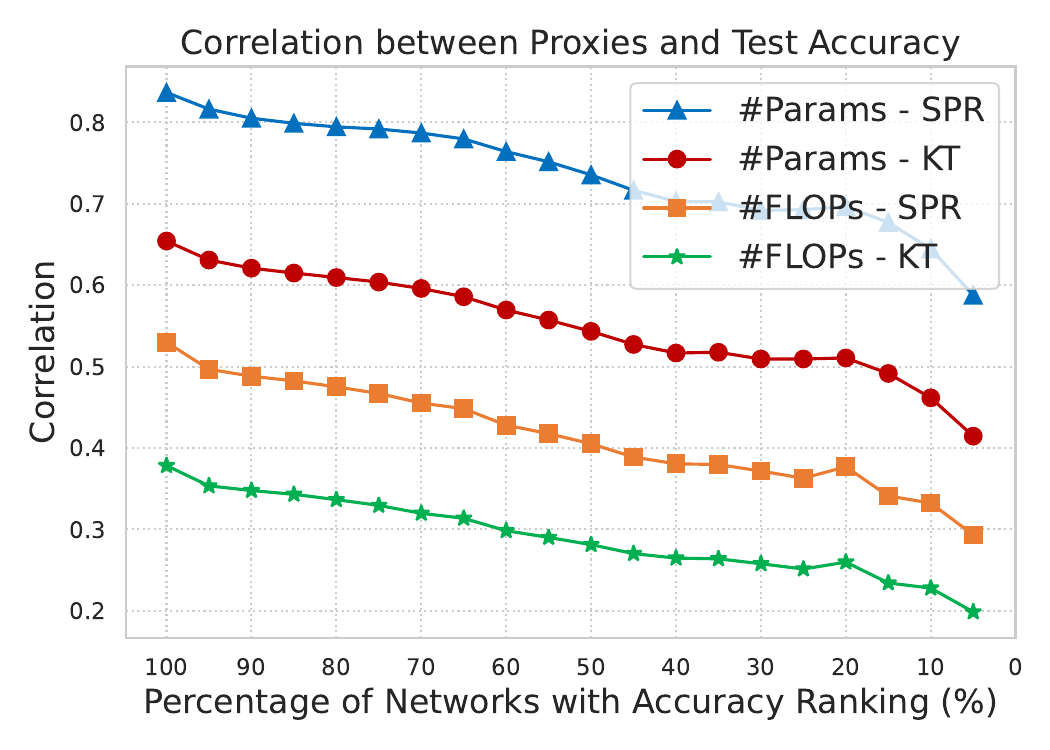}
 \caption{The correlation between \#Params \& \#FLOPs and the test accuracy under various ratios of networks on NATS-Bench for ImageNet16-120 dataset (averaged over 5 seeds).}
 \label{fig:ratio_nats_img}
\end{figure}

\begin{table*}[t]
\centering
 \caption{\color{black} Comparison of zero-shot proxies based NAS vs. one-shot NAS on ProxylessNAS search space. The results are averaged over three runs. }\label{tab:os_vs_zs}
 \label{tab:my_label}
 \centering
\scalebox{0.95}{\color{black}
 \begin{tabular}{c|c|c|c|c|c|c|c|c|c}
 \toprule
 \textbf{Method} & \textbf{One-shot NAS} & \textbf{Grad\_norm} & \textbf{Synflow} & \textbf{GradSign} & \textbf{Jacob\_cov} & \textbf{NTK\_Cond} & \textbf{Zen-score} & \textbf{Params} & \textbf{FLOPs} \\ \midrule
 \textbf{Top-1 on ImageNet-1K} & 74.39 & 71.46 & 70.02 & 73.17 & 70.31 & 73.63 & 71.78 & 72.87 & 73.08 \\ \midrule
 \textbf{mAP on COCO} & 0.28 & 0.22 & 0.21 & 0.23 & 0.24 & 0.27 & 0.25 & 0.26 & 0.28 \\ \midrule
 \textbf{Search cost (GPU Hours)} & 200 & 8.9 & 8.8 & 9.7 & 9.2 & 37 & 1.6 & 0.03 & 1.5 \\ \bottomrule
 \end{tabular}
}
\end{table*}

\begin{figure}
 \centering
 \includegraphics[width=0.45\textwidth]{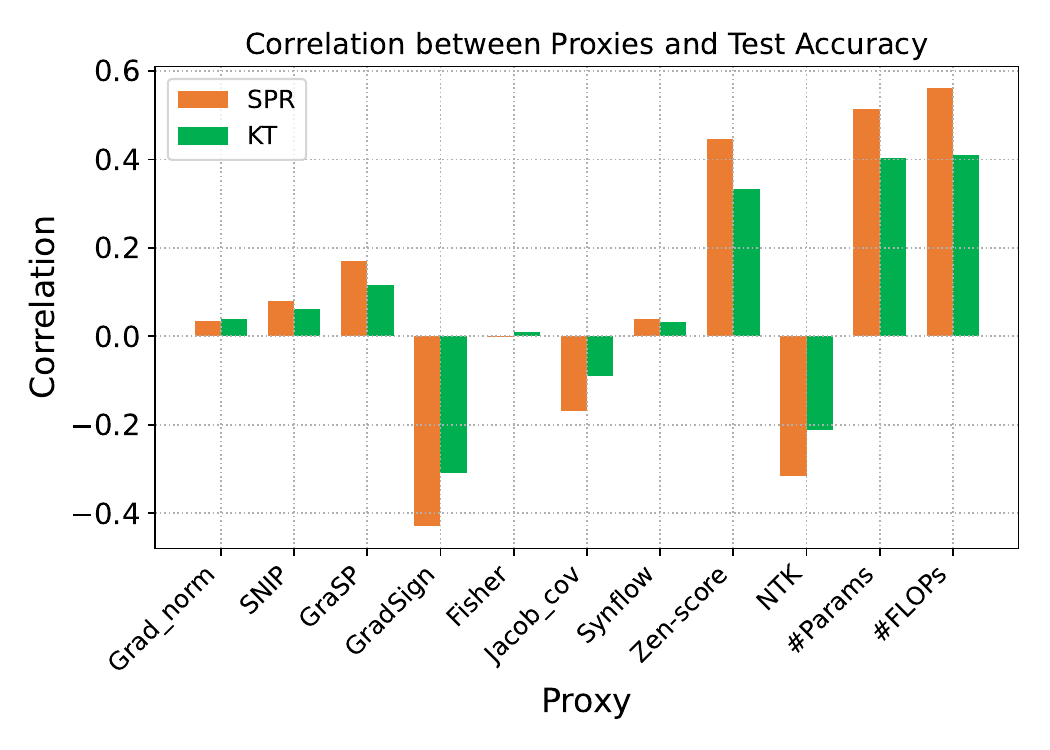}
 \caption{{\color{black}The correlation between various proxies and the test accuracy on the CNNs model space for ImageNet-1K classification.}}
 \label{fig:timm_cnn}
\end{figure}

\begin{figure}
 \centering
 \begin{subfigure}[b]{0.45\textwidth}
 \centering
 \includegraphics[width=\textwidth]{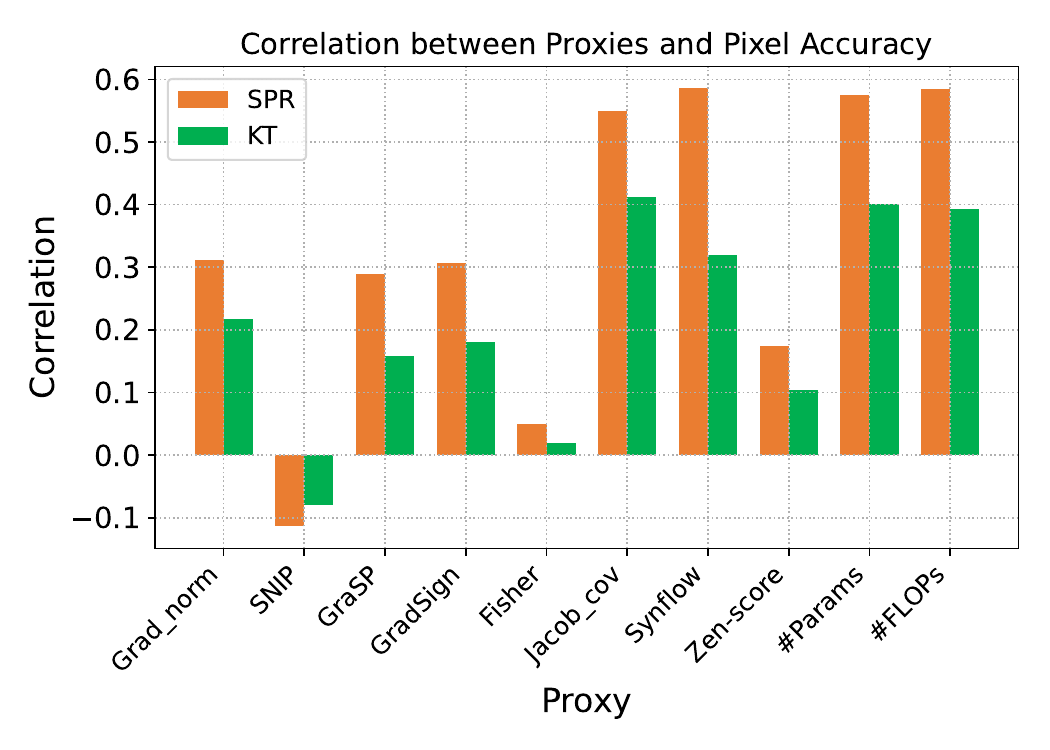}\vspace{-3.5mm}
 \caption{\color{black} Pixel Accuracy}
 \label{fig:ade20k_acc}
 \end{subfigure}
 
 \begin{subfigure}[b]{0.45\textwidth}
 \centering
 \includegraphics[width=\textwidth]{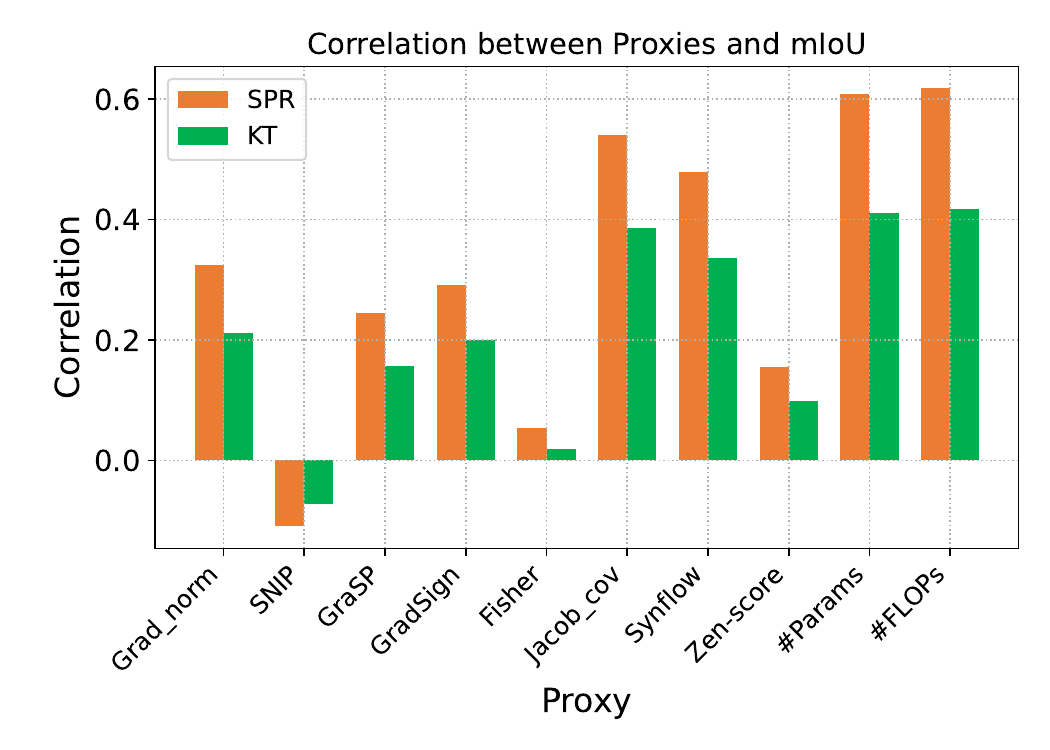}\vspace{-3.5mm}
 \caption{\color{black} mIoU}
 \label{fig:ade20k_miou}
 \end{subfigure}
 \caption{{\color{black}The correlation between various proxies and pixel accuracy (or mIoU) on ADE20K semantic segmentation.}}
 \label{fig:ade20k}
\end{figure}

\begin{figure}
 \centering
 \includegraphics[width=0.45\textwidth]{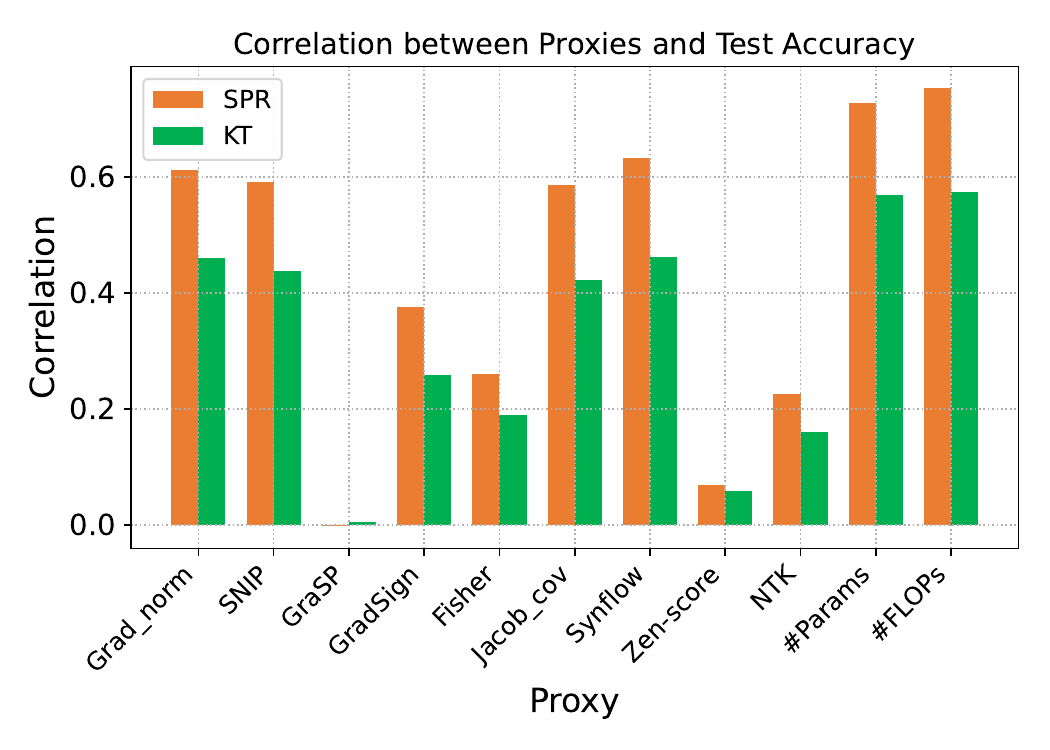}
 \caption{{\color{black}The correlation between various proxies and the test accuracy on the ViT model space for ImageNet-1K classification.}}
 \label{fig:timm_vit}
\end{figure}
As shown in previous literature, \#Params and \#FLOPs outperform other proxies in multiple benchmarks~\cite{zeroshotnassurvery}. Hence, we dig deep into the effectiveness of \#Params and \#FLOPs by gradually making the search space more constrained. As shown in Figure~\ref{fig:ratio_201_c100} and Figure~\ref{fig:ratio_nats_img}, if we compute the correlation for networks with higher accuracy, both \#Params and \#FLOPs have a significant drop in correlation score. 

Given the above results, we conclude that all of the existing proxies (including \#Params and \#FLOPs) do \textit{not} correlate well for the network with high accuracy. This is a fundamental drawback because what matters most for NAS are precisely these networks with high accuracy. Hence, there is great potential for designing better proxies that could yield high correlation scores for these top networks.

\subsubsection{Specific Network Families}\label{sec:compare_nnmass}
We remark that many popular neural architectures are not included in most NAS benchmarks. Hence, in this section, we consider several commonly used network families as the search space since they are widely used in various applications.
As shown in Figure~\ref{fig:nnmass_resnet}, if we search within networks from ResNet and Wide-ResNet families, then SNIP, Zen-score, \#Params, \#FLOPs, and NN-Mass have a significantly high correlation with the test accuracy (\textit{i.e.,} Spearman's $\rho>0.9$). 

As shown in Figure~\ref{fig:nnmass_mbnv2}, Grad\_norm, SNIP, Fisher, Synflow, Zen-score, and NN-Mass work best for the MobileNet-v2 network family, which is slightly better than the two naive proxies \#Params and \#FLOPs. These results show that there is great potential in designing good proxies for a constrained yet widely used search space.

\begin{figure*}
 \centering
 \begin{subfigure}[b]{0.32\textwidth}
 \centering
 \includegraphics[width=\textwidth]{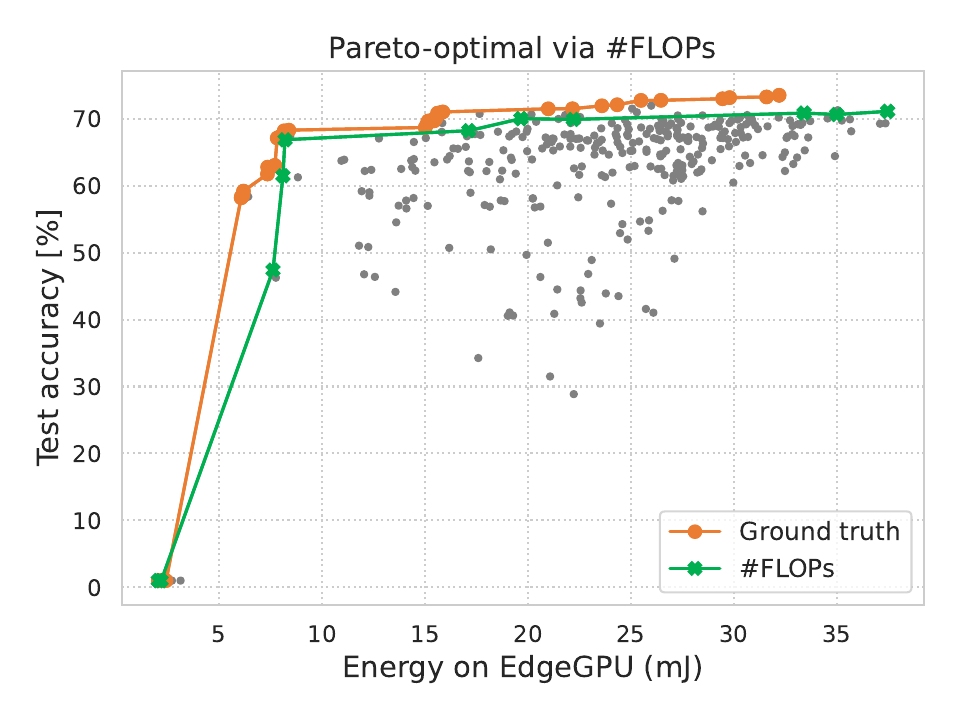}\vspace{-2mm}
 \caption{\#FLOPs}
 \end{subfigure}
 \hfill
 \begin{subfigure}[b]{0.32\textwidth}
 \centering
 \includegraphics[width=\textwidth]{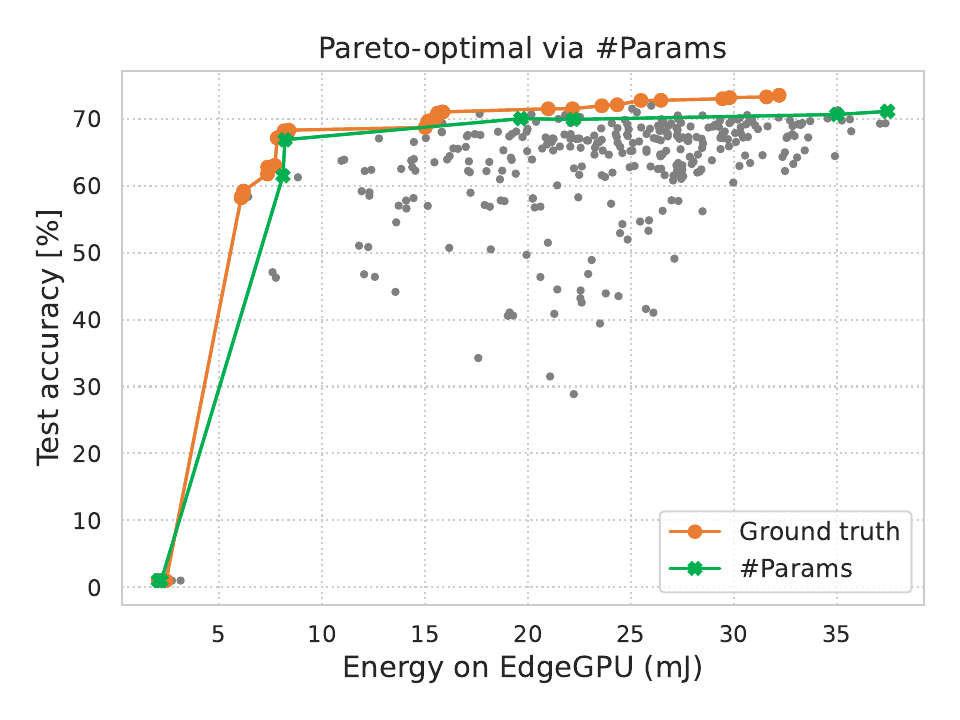}\vspace{-2mm}
 \caption{\#Params}
 \end{subfigure}
 \hfill
 \begin{subfigure}[b]{0.32\textwidth}
 \centering
 \includegraphics[width=\textwidth]{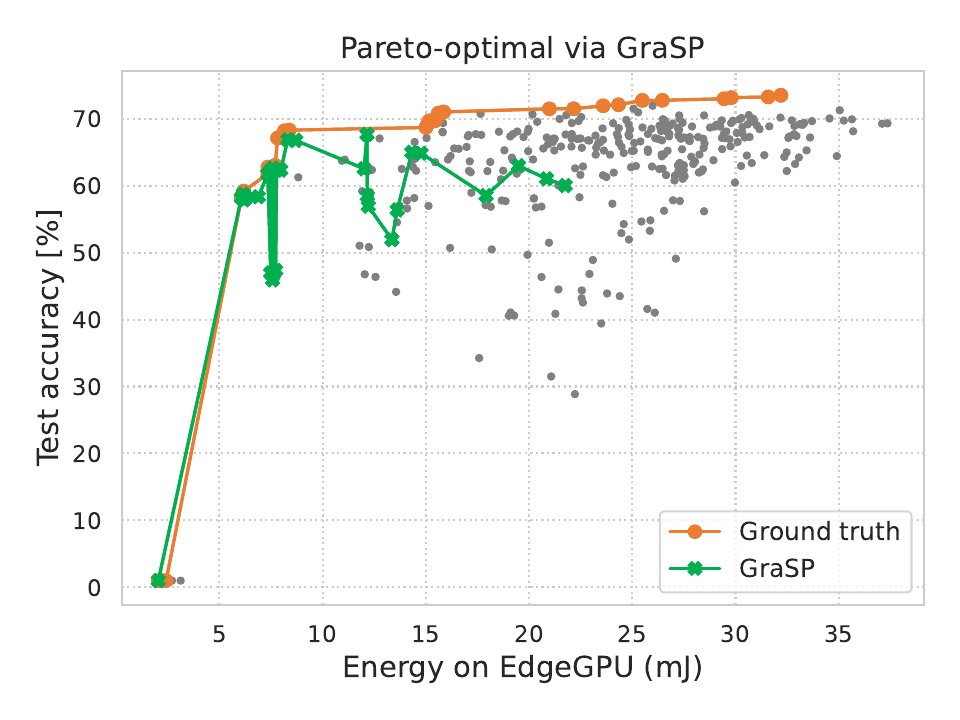}\vspace{-2mm}
 \caption{GraSP}
 \end{subfigure}

 \begin{subfigure}[b]{0.32\textwidth}
 \centering
 \includegraphics[width=\textwidth]{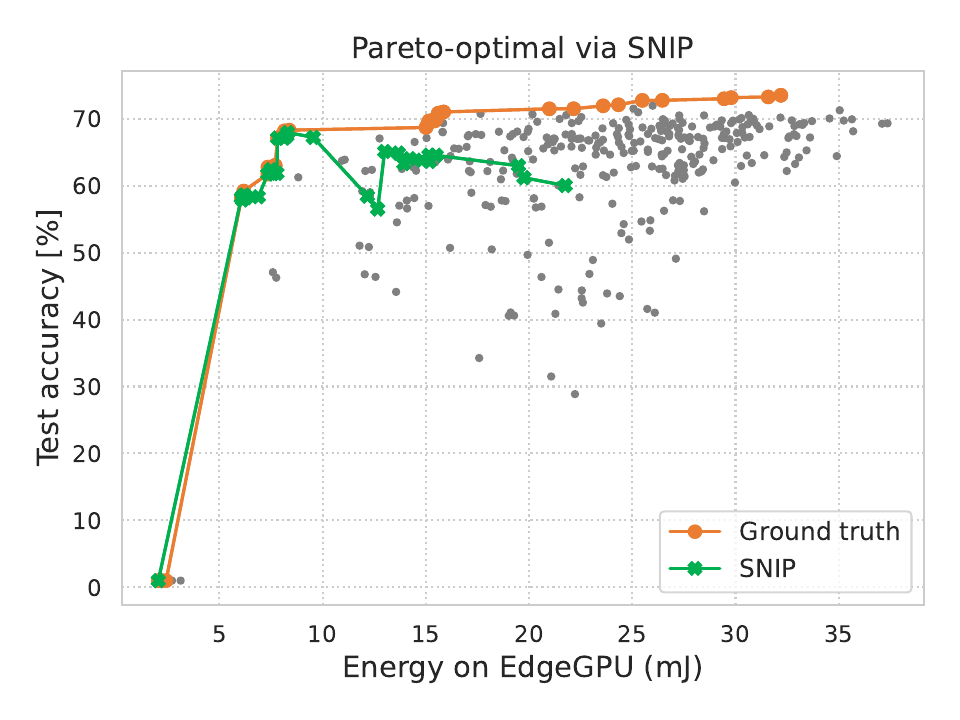}\vspace{-2mm}
 \caption{SNIP}
 \end{subfigure}
 \hfill
 \begin{subfigure}[b]{0.32\textwidth}
 \centering
 \includegraphics[width=\textwidth]{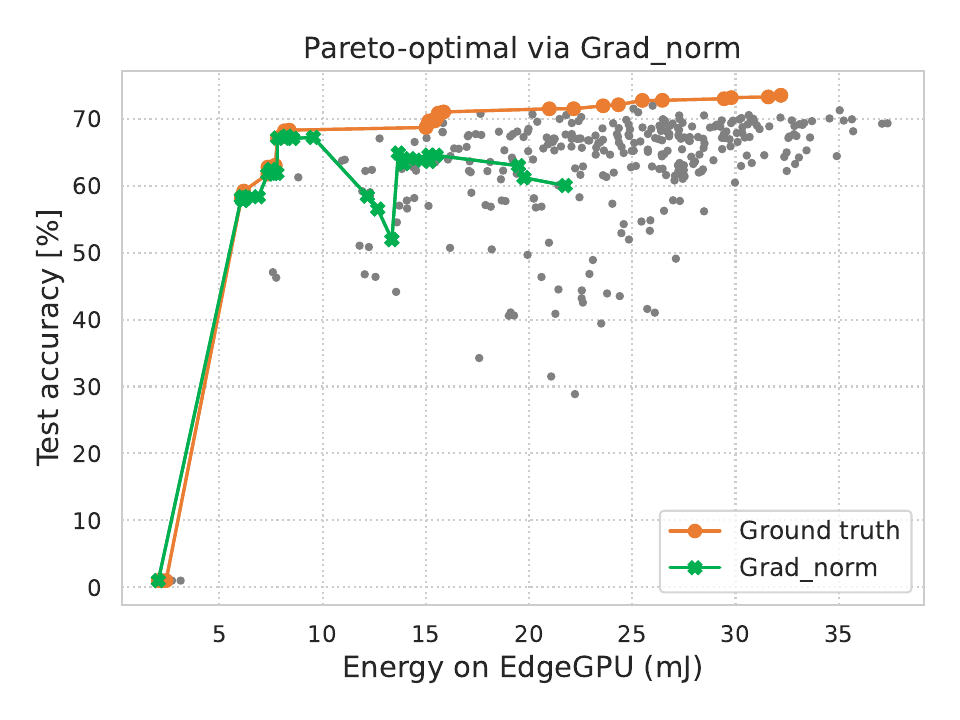}\vspace{-2mm}
 \caption{Grad\_norm}
 \end{subfigure}
 \hfill
 \begin{subfigure}[b]{0.32\textwidth}
 \centering
 \includegraphics[width=\textwidth]{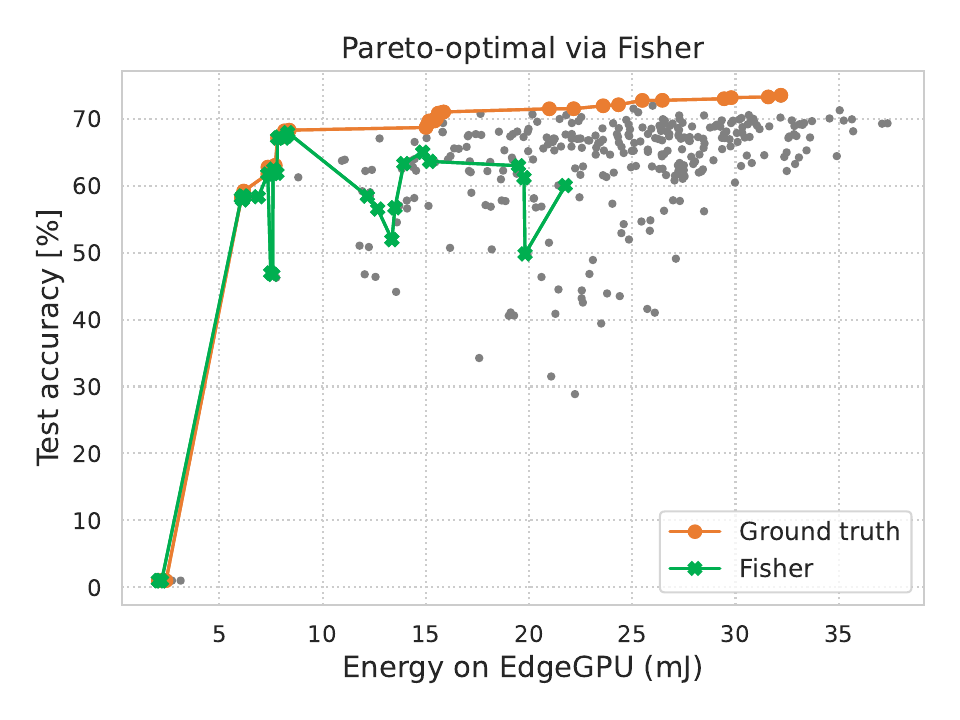}\vspace{-2mm}
 \caption{Fisher information}
 \end{subfigure}

 \begin{subfigure}[b]{0.32\textwidth}
 \centering
 \includegraphics[width=\textwidth]{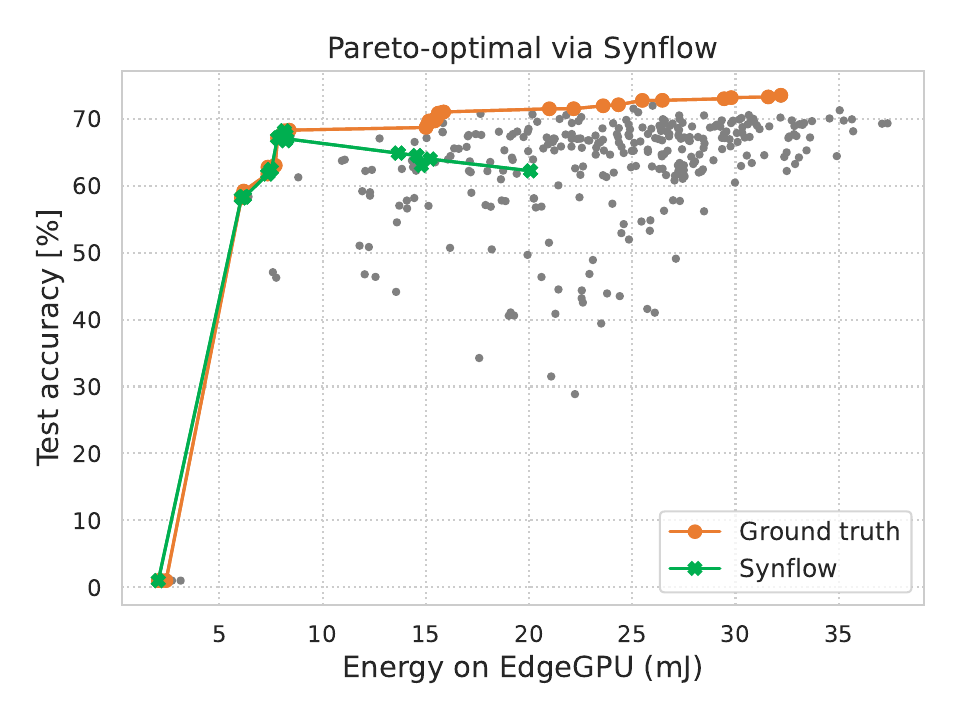}\vspace{-2mm}
 \caption{Synflow}
 \end{subfigure}
 \hfill
 \begin{subfigure}[b]{0.32\textwidth}
 \centering
 \includegraphics[width=\textwidth]{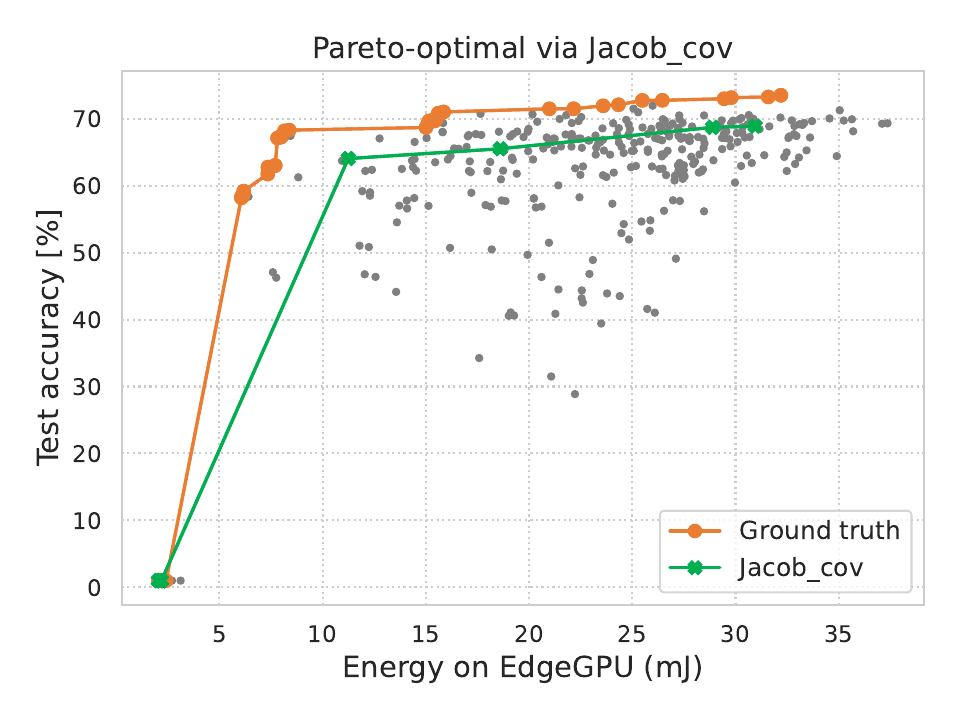}\vspace{-2mm}
 \caption{Jacob\_cov}
 \end{subfigure}
 \hfill
 \begin{subfigure}[b]{0.32\textwidth}
 \centering
 \includegraphics[width=\textwidth]{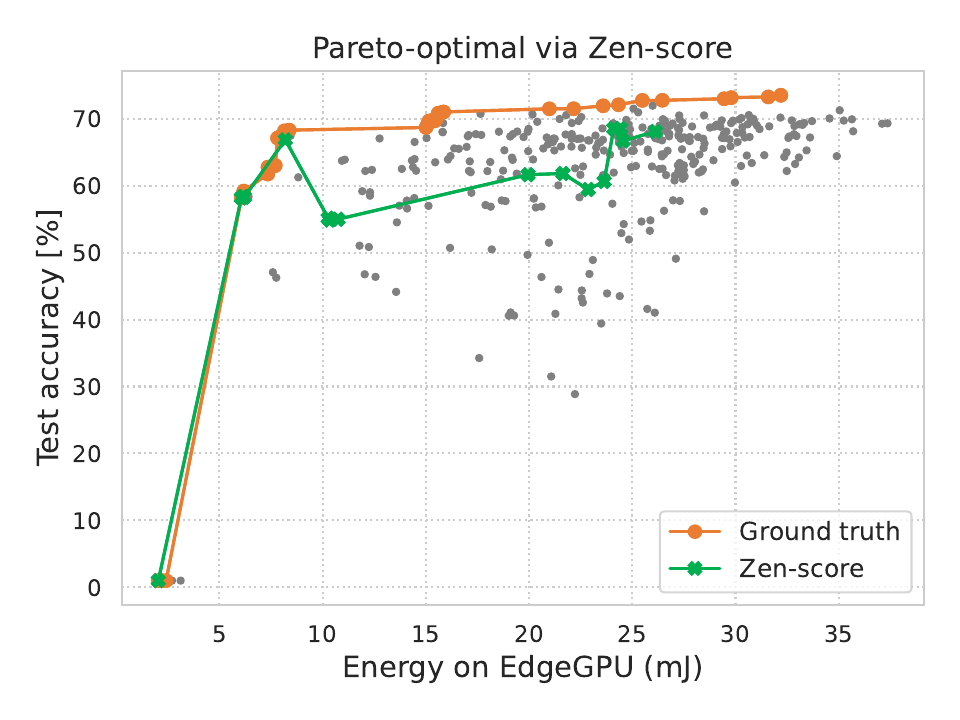}\vspace{-2mm}
 \caption{Zen-score}
 \end{subfigure}
 \caption{Pareto-optimal networks obtained via various proxies for CIFAR100 dataset on NASBench-201, and for various energy consumption constraints on an EdgeGPU (NVIDIA Jetson TX2). The gray points in these figures are candidate networks in the search space. }
 \label{fig:pareto_201_c100_energy_edgegpu}
\end{figure*}

\begin{figure*}
 \centering
 \begin{subfigure}[b]{0.32\textwidth}
 \centering
 \includegraphics[width=\textwidth]{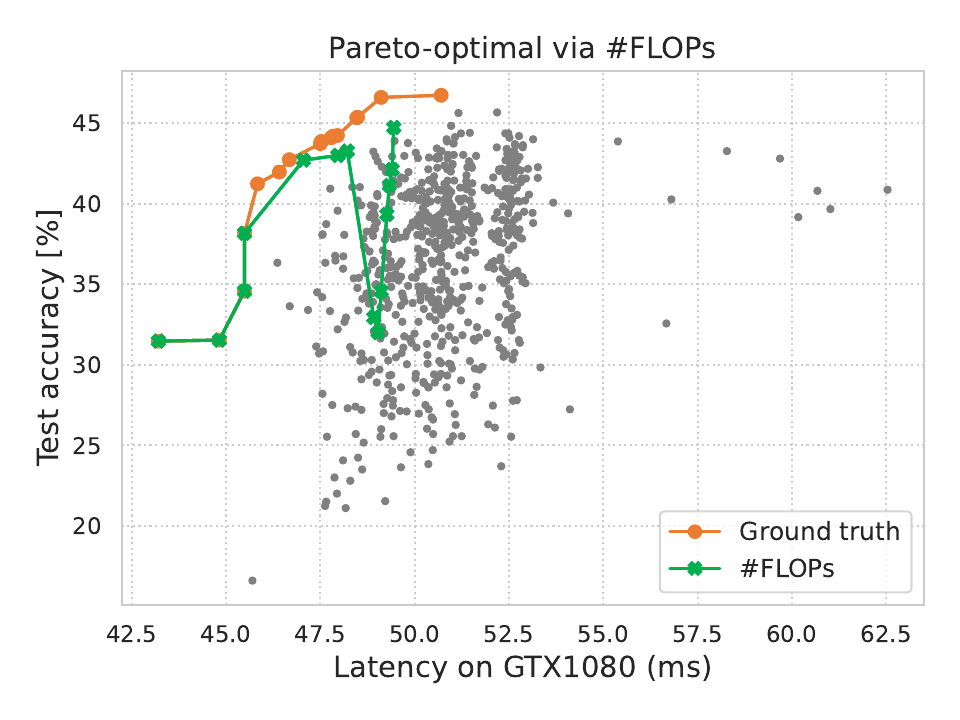}\vspace{-2mm}
 \caption{\#FLOPs}
 \end{subfigure}
 \hfill
 \begin{subfigure}[b]{0.32\textwidth}
 \centering
 \includegraphics[width=\textwidth]{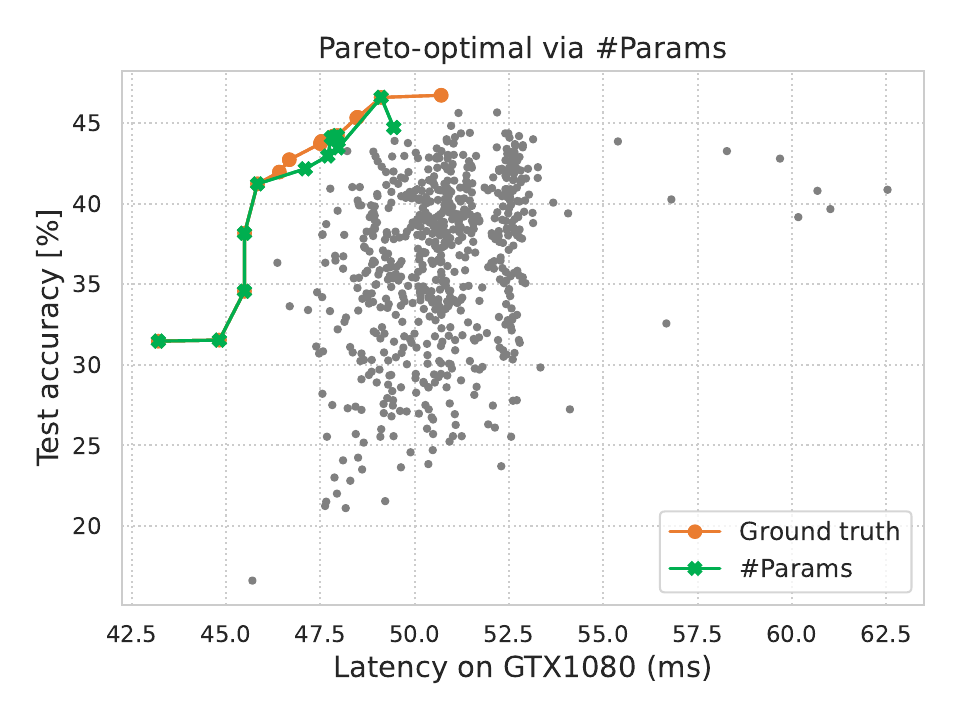}\vspace{-2mm}
 \caption{\#Params}
 \end{subfigure}
 \hfill
 \begin{subfigure}[b]{0.32\textwidth}
 \centering
 \includegraphics[width=\textwidth]{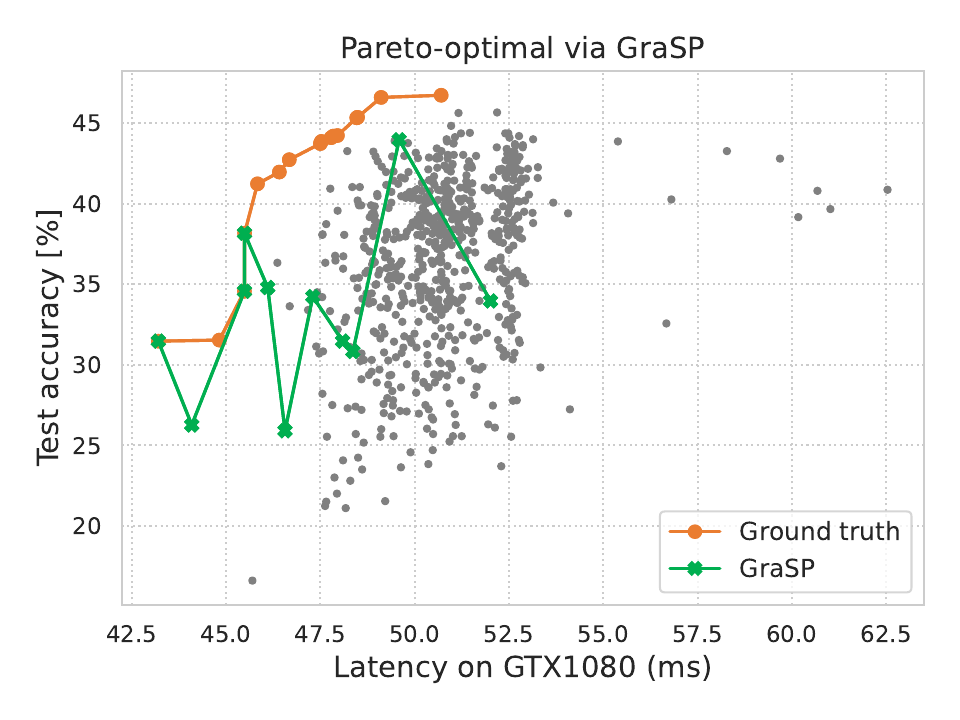}\vspace{-2mm}
 \caption{GraSP}
 \end{subfigure}

 \begin{subfigure}[b]{0.32\textwidth}
 \centering
 \includegraphics[width=\textwidth]{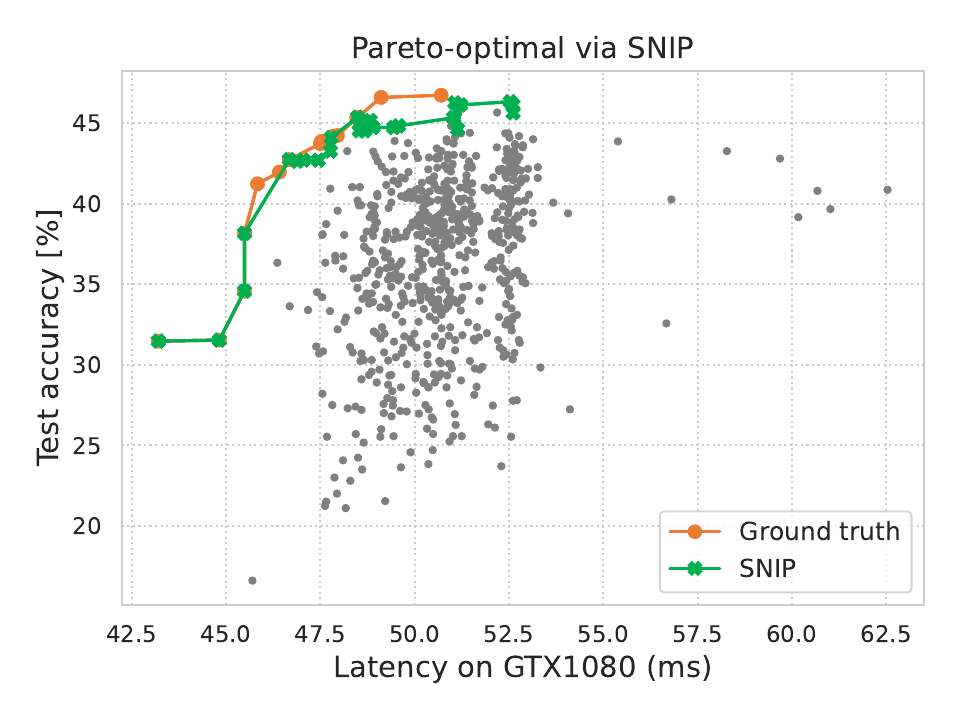}\vspace{-2mm}
 \caption{SNIP}
 \end{subfigure}
 \hfill
 \begin{subfigure}[b]{0.32\textwidth}
 \centering
 \includegraphics[width=\textwidth]{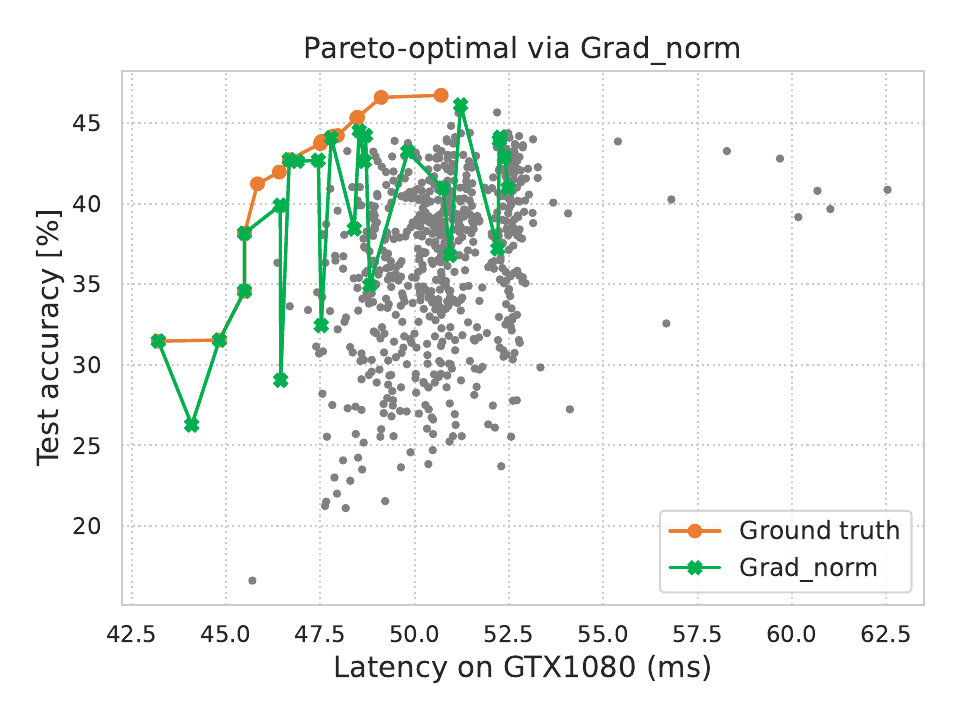}\vspace{-2mm}
 \caption{Grad\_norm}
 \end{subfigure}
 \hfill
 \begin{subfigure}[b]{0.32\textwidth}
 \centering
 \includegraphics[width=\textwidth]{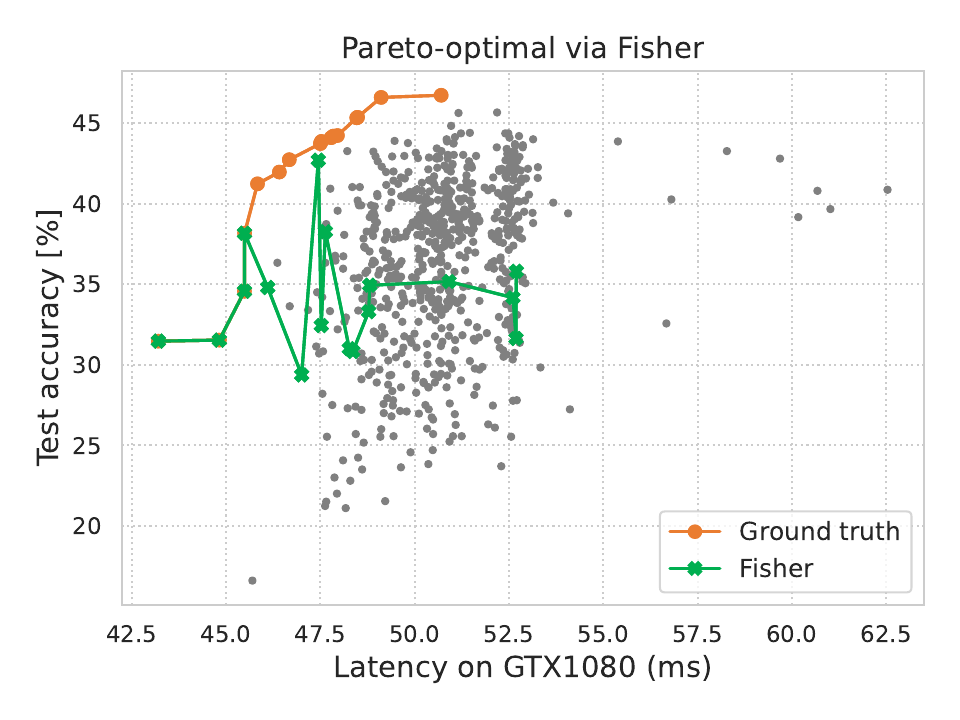}\vspace{-2mm}
 \caption{Fisher information}
 \end{subfigure}

 \begin{subfigure}[b]{0.32\textwidth}
 \centering
 \includegraphics[width=\textwidth]{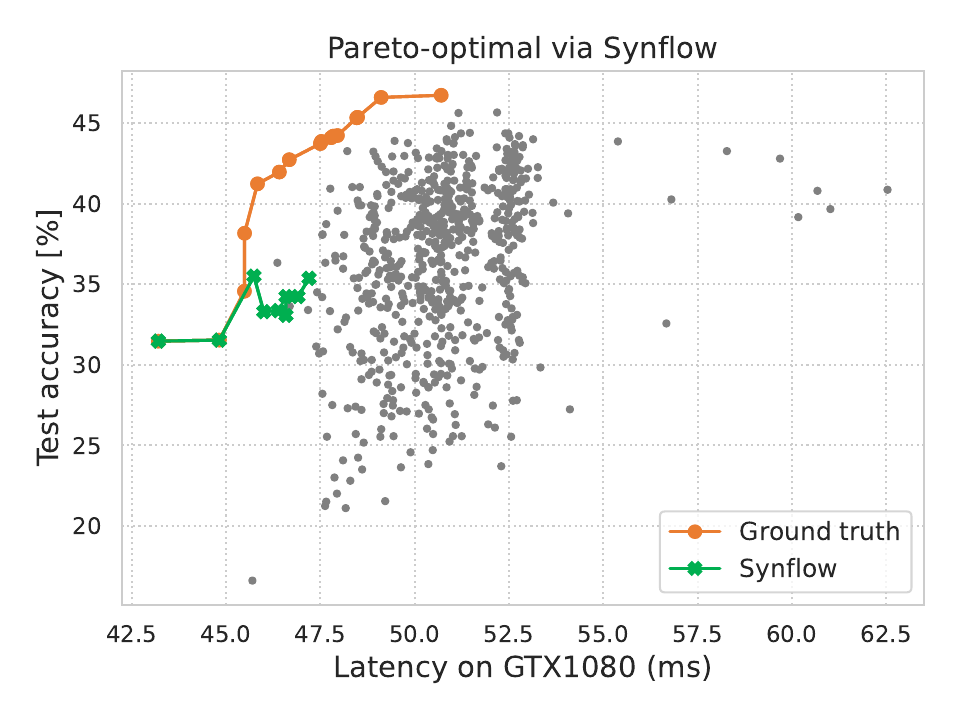}\vspace{-2mm}
 \caption{Synflow}
 \end{subfigure}
 \hfill
 \begin{subfigure}[b]{0.32\textwidth}
 \centering
 \includegraphics[width=\textwidth]{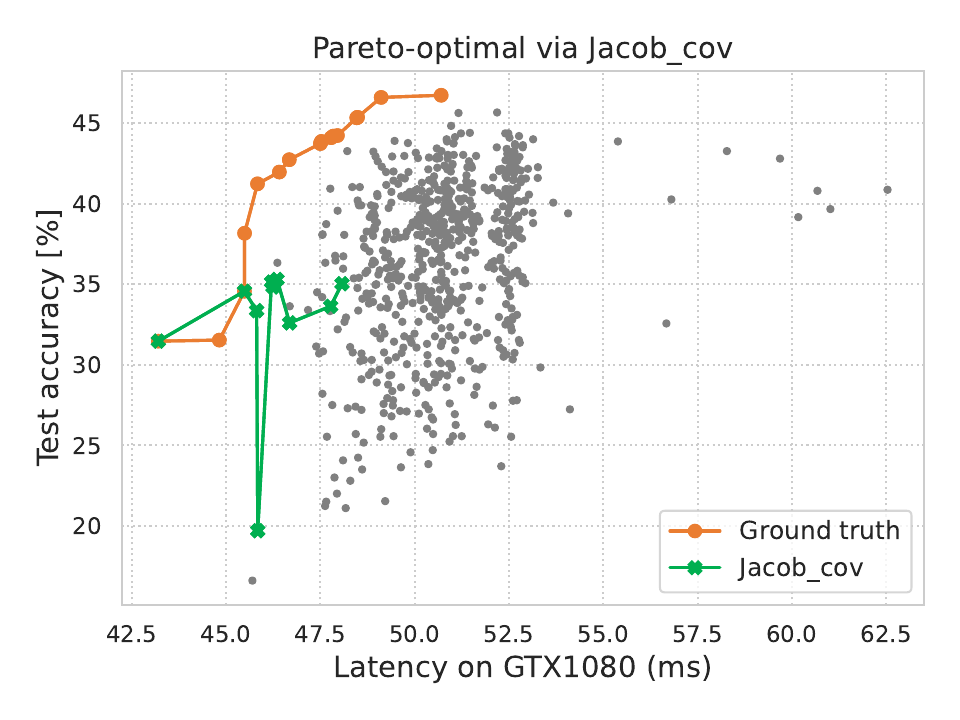}\vspace{-2mm}
 \caption{Jacob\_cov}
 \end{subfigure}
 \hfill
 \begin{subfigure}[b]{0.32\textwidth}
 \centering
 \includegraphics[width=\textwidth]{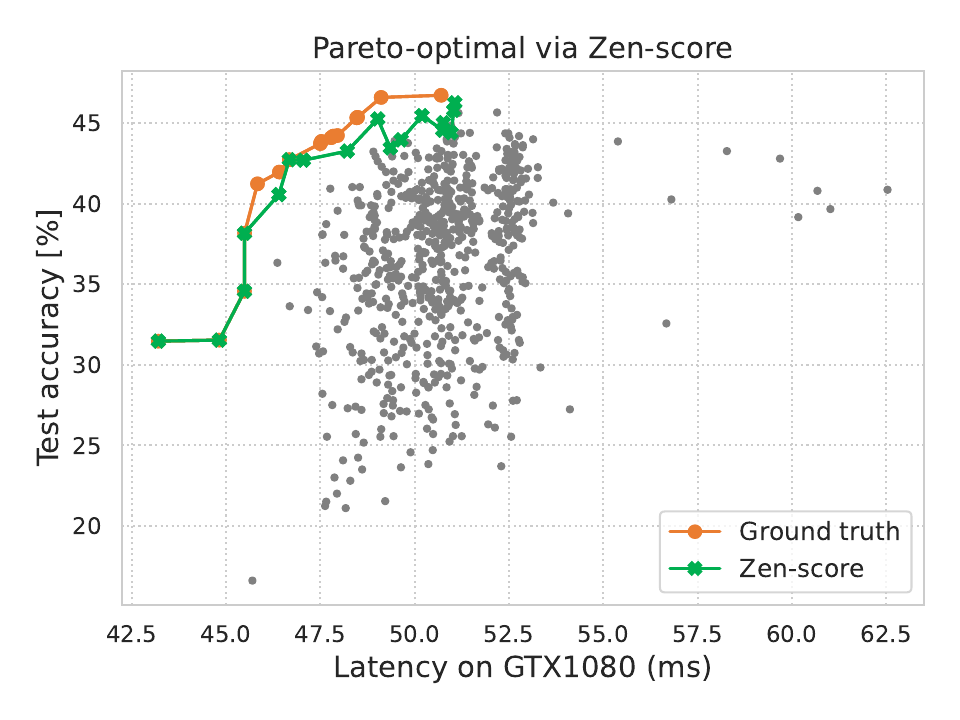}\vspace{-2mm}
 \caption{Zen-score}
 \end{subfigure}
 \caption{Pareto-optimal networks obtained via various proxies for ImageNet16-120 dataset on NATS-Bench, and for various latency constraints on NVIDIA GTX1080. The gray points in these figures are candidate networks in the search space. }
 \label{fig:pareto_nats_img_lat_1080}
\end{figure*}

{\color{black}
\subsection{Large-scale Dataset}

To further compare these proxies in more complicated scenarios, we illustrate the performance for ImageNet-1K classification, COCO object detection, and ADE20K semantic segmentation tasks.

\noindent\textbf{ImageNet-1K classification.}
We first compute the zero-shot proxies for the CNN architectures in the model space of TIMM~\cite{rw2019timm}. Notably, we only consider networks that are trained standalone on ImageNet-1K without pre-training or distillation. In total, we evaluate 200 CNNs and report the correlation between Top-1 accuracy and multiple proxies in Figure~\ref{fig:timm_cnn}. As shown, \#Params and \#FLOPs still have a higher correlation than these zero-shot proxies. This is consistent with our observations on NAS benchmarks.

We also compare the performance of these proxy-based NAS with one-shot NAS within the same search space. We conduct the comparison on the MobileNet-V2 based search space under the same \#FLOPs budget of 600M. Specifically, for proxy-based NAS, we use the evolutionary algorithm to search for the architecture with the highest proxy values; we conduct the search for at most 10K steps. For the one-shot NAS, we use the same algorithm from~\cite{cai2018proxylessnas}. We train the obtained architecture for 150 epochs under the standard data augmentation configurations. We use the SGD optimizer with an initial learning rate of 0.1 and a cosine annealing learning rate schedule.

As shown in Table~\ref{tab:os_vs_zs}, compared to one-shot NAS, zero-shot proxy-based NAS has a slight accuracy degradation (less than 1\%), but requires orders of magnitude less search costs. Moreover, when comparing these zero-shot proxies, NTK\_Cond based search performs closest to one-shot NAS, but at a higher search cost than other proxies. These results highlight an intrinsic trade-off between search cost and the accuracy of the obtained architectures.

\noindent\textbf{COCO object detection.}
Following the standard practice in NAS, we employ the architectures obtained on ImageNet-1K (shown in Table~\ref{tab:os_vs_zs}) as the backbone for detection models. By using the detection head from NanoDet~\cite{nanodet}, we then train these networks for 50 epochs on COCO following the same training setup as NanoDet.
As shown in Table~\ref{tab:os_vs_zs}, the results follow a trend similar to that of ImageNet-1K. More precisely, \#FLOPs and NTK\_Cond based zero-shot NAS yield performance that is the same or very close to the one-shot NAS.

\noindent\textbf{ADE20K semantic segmentation.} We compute these zero-shot proxies for the CNN architecture in the model space of PyTorch Segmentation~\cite{rw2019timm}. We vary both the backbone and segmentation heads to obtain multiple segmentation networks; we then train these models from scratch and get their test performance. In total, we evaluate 200 CNNs and report the correlation between pixel accuracy (or mIoU) and various proxies in Figure~\ref{fig:ade20k}. As shown, \#Params and \#FLOPs have a higher correlation than the other zero-shot proxies. 

To conclude, these comprehensive evaluations on these large-scale datasets reaffirm the dominance of \#Params and \#FLOPs over other proxies in multiple scenarios. Therefore, future works should make comprehensive comparisons under various tasks and datasets to show a consistent advantage over \#Params and \#FLOPs. Besides, while zero-shot proxy-based NAS exhibits certain efficiencies, there remains a trade-off between search cost and test performance accuracy.

\subsection{Vision Transformers}
Until now, our evaluations have primarily focused on CNNs; however, with the recent surge in their performance and popularity, vision transformers (ViTs) are becoming increasingly important in the realm of computer vision~\cite{dosovitskiy2021an_vit}. Therefore, in this section, we evaluate these proxies using the ViT model space for ImageNet-1K.

Specifically, we compare these zero-shot proxies for the ViTs in the model space of TIMM~\cite{rw2019timm}. Notably, we only include networks that are trained standalone on ImageNet-1K without pre-training or distillation. In total, we evaluate 100 ViTs and report the correlation between Top-1 accuracy and various proxies in Figure~\ref{fig:timm_vit}. The results show that \#Params and \#FLOPs has higher correlation score with the test accuracy than the zero-shot proxies. This is consistent with our observations on CNNs. 
In conclusion, whether analyzing CNNs or ViTs, the superior correlation of \#Params and \#FLOPs over zero-shot proxies is consistent. 

In practical applications, test performance is not the only design consideration. Indeed, the models obtained by NAS should meet some hardware constraints, especially for deployment on edge devices. Hence, we next explore the performance of these proxies for the hardware-aware search scenarios. 

}

\subsection{Hardware-aware NAS}\label{sec:comparison_hwnas}

In this part, we conduct the hardware-aware NAS using the zero-shot proxies introduced above. Specifically, we use these zero-shot proxies instead of the real test accuracy to search for the Pareto-optimal networks under various constraints. We next introduce the results on NASBench-201 (with HW-NAS-Bench) and NATS-Bench.

\subsubsection{NASBench-201 / HW-NAS-Bench}
We use EdgeGPU (NVIDIA Jetson TX2) as the target hardware and use the energy consumption data from HW-NAS-Bench; then we set various energy consumption values as the hardware constraints. Next, we use different accuracy proxies to traverse all candidate architectures in the search space and obtain the Pareto-optimal networks under various energy constraints. 

To illustrate the quality of these networks, we plot these networks and the ground truth results obtained via actual accuracy in Figure~\ref{fig:pareto_201_c100_energy_edgegpu}. As shown, when the energy constraint is tight (\textit{e.g.,} less than 10mJ), most of the proxies could find networks very close to the real Pareto-optimal, except the Jacob\_cov. However, when the energy constraint is more relaxed (\textit{e.g.,} more than 20mJ), only \#Params, \#FLOPs, and Jacob\_cov can find several networks close to the ground truth. 

\subsubsection{NATS-Bench}
We measure the latency data on NVIDIA GTX-1080 for NATS-Bench. We then use different accuracy proxies to traverse all candidate architectures to obtain the Pareto-optimal networks under various latency constraints. As shown in Figure~\ref{fig:pareto_nats_img_lat_1080}, we plot these networks and the ground truth results. When we set the latency constraint to around 50ms, only \#Params, SNIP, and Zen-score can still find the networks that nearly match the real Pareto-optimal networks. 

The results on these two benchmarks further verify that current proxies don't correlate well for networks with high accuracy because the real Pareto-optimal networks have higher accuracy when the hardware constraints are more relaxed. This observation suggests a great potential to design better proxies in this scenario. 

\subsection{Discussion and future work}

\subsubsection{NAS Benchmarks}
\noindent\textit{Diversity of search space:}
We remark that the search space of most existing NAS benchmarks only contains cell-based neural architectures. To further improve the generality of NAS benchmarks, the community may need to incorporate new architectures from more diverse search spaces. For instance, the NATS-Bench has added architectures with different cells for different stages of the search space. 
Moreover, the cells in these existing benchmarks are similar to the DARTS cell structure. However, in practice, the inverted bottleneck blocks from MobileNet-v2 are more widely used for higher hardware efficiency. Therefore, the next direction of NAS benchmarks may need to cover a more practical and widely used search space, such as FBNet-v3.

\noindent\textit{Awareness of hardware efficiency:}
So far, only HW-NAS-Bench provides multiple hardware constraints on several types of hardware platforms, but it does not have the accuracy data for most of the networks in the benchmark. Thus, we recommend future NAS benchmarks to incorporate both accuracy and hardware metrics for typical hardware platforms.

\subsubsection{Zero-shot proxies}

\noindent\textit{Why \#Params works:}
As shown in Section~\ref{sec:exp_unconstrained}, \#Params achieves a higher correlation than other proxies with multiple datasets and multiple benchmarks for unconstrained search space. One may wonder why such a trivial proxy works so well. In general, a good neural architecture should satisfy the following properties: good convergence/trainability and high expressive capacity.
We provide the following observations:
\begin{itemize}
 \item \textbf{Expressive Capacity}\quad {\color{black} It is well known that a network with infinite width or depth, can express any type of complex functions with an arbitrarily small errors~\cite{universal_dedth,universal_width,tripuraneni2021overparameterization}. Moreover, previous works show that, with the depth or width values increasing, the error w.r.t. ground truth functions will gradually decrease. In other words, more parameters capture the higher expressive capacity of a given neural network~\cite{express_survey}.}
 \item \textbf{Generalization Capacity}\quad Previous work reveals that a network with more parameters tends to have higher test accuracy under an appropriate training setup~\cite{Overparameterized}.
 \item \textbf{Trainability}\quad On the one hand, given similar depth, the wider networks have better trainability and higher convergence rates, and clearly more parameters~\cite{nnmass}. On the other hand, most of the networks evaluated on popular benchmarks share a similar depth value. Hence, within these benchmarks, more parameters will also indicate a better trainability.
\end{itemize}
Hence, \#Params captures both the expressivity and trainability of the networks in these benchmarks. 
In contrast, most of the proposed proxies usually emphasize either the expressivity or the trainability of networks (but not both). That may be why \#Params outperforms these proposed proxies. Hence, future work should aim to design a proxy that could indicate both the convergence/trainability and expressive and generalization capacity of a given network. For instance, recently proposed proxy ZiCo indicates both trainability and generalization capacity of neural networks thus consistently outperforming \#Params in multiple NAS benchmarks~\cite{li2023zico}. 

\noindent\textit{When \#Params fails:} (\textit{i}) As shown in this section, when accounting for the architectures with test accuracy ranking top 5\%, several proxies outperform both \#Params and \#FLOPs for some benchmarks. Furthermore, these top-performing network architectures are most important since NAS focuses on obtaining the networks with high accuracy. (\textit{ii}) Many proxies work well in the constrained search space, such as the MobileNet and ResNet families. These networks are widely used in many applications (\textit{e.g.,} MobileNet-v2 for EdgeAI). Clearly, the above two failing cases are very important to push zero-shot NAS to more practical scenarios. Hence, there is a great potential to explore better zero-shot proxies in the above cases. 

\noindent\textit{Search method:}
Though \#Params outperforms most proxies in several scenarios in terms of correlation coefficients, there are alternative search methods to use these zero-shot proxies. For example, as demonstrated in ~\cite{tf_nas}, to better leverage these proxies, one potential search method can merge all candidate networks into a supernet and then apply these proxies to prune the network at the initialization stage until hardware constraints are met. This way, the time efficiency of zero-shot NAS approaches can be further improved since the search space is gradually compressed with pruning going on.

\noindent\textit{Theoretical support:}
We remark that most gradient-based proxies are first proposed to estimate the importance of each parameter or neuron/channel of a given network, thus originally applied to the model pruning problem space instead of ranking networks. Hence, the effectiveness of these gradient-based proxies for zero-shot NAS needs a more profound understanding from a theoretical perspective. Moreover, though most gradient-free proxies are usually presented with some theoretical analysis for NAS, as shown in Section~\ref{sec:comparison_purenas} and Section~\ref{sec:comparison_hwnas}, they generally have a lower correlation with the gradient-based ones. The theoretical understanding of why these zero-shot proxies can or cannot estimate the test accuracy of different networks is still an open question.

\noindent\textit{Customized proxy for different types of networks:}
As mentioned in Section~\ref{sec:compare_nnmass}, several zero-shot proxies do not work well for a general search space, but do show a great correlation with the test accuracy and beat the \#Params on constrained search spaces. In fact, Section~\ref{sec:comparison_purenas} and Section~\ref{sec:comparison_hwnas} show that designing a zero-shot proxy that generally works well is extremely difficult. One potential direction for the design of zero-shot proxies may lie in partitioning the entire search space into several sub-spaces and then proposing customized proxies specifically designed for different sub-spaces.

\section{Conclusion}\label{sec:conclusion}
In this paper, we have presented a comprehensive review of existing zero-shot NAS approaches. To this end, we have first introduced accuracy proxies for zero-shot NAS by providing theoretical inspirations behind these proxies, and several commonly used NAS benchmarks. We then have introduced several popular approaches for hardware performance predictions. We have also compared the existing proxies against two naive proxies, namely, \#Params and \#FLOPs. By calculating the correlation between these proxies and the real test accuracy, we have shown that the proposed proxies to date are not necessarily better than \#Params and \#FLOPs for these tasks for unconstrained search spaces (\textit{i.e.,} considering all architectures in benchmarks). However, for constrained search spaces (\textit{i.e.,} when considering only networks with high accuracy), we have revealed that the existing proxies, including \#Params and \#FLOPs, has much worse correlation scores with the real accuracy than unconstrained scenarios. Based on these analyses, we have explained why \#Params work and when \#Params fail. Finally, we have pointed out several potential research directions to design better benchmarks for better zero-shot NAS and multiple ideas that may enable the design of better zero-shot NAS approaches.

\appendices


\section*{Acknowledgments}
Radu Marculescu and Guihong Li are supported in part by the NSF grant CNS 2007284, and in part by the iMAGiNE Consortium [\href{https://imagine.utexas.edu/}{Link}].
Z. Wang is in part supported by NSF Scale-MoDL (\#2133861).

\ifCLASSOPTIONcaptionsoff
 \newpage
\fi



%

\bibliographystyle{IEEEtran}
\bibliography{reference}



%

\begin{IEEEbiography}[{\includegraphics[width=1in,height=1.25in,clip,keepaspectratio]{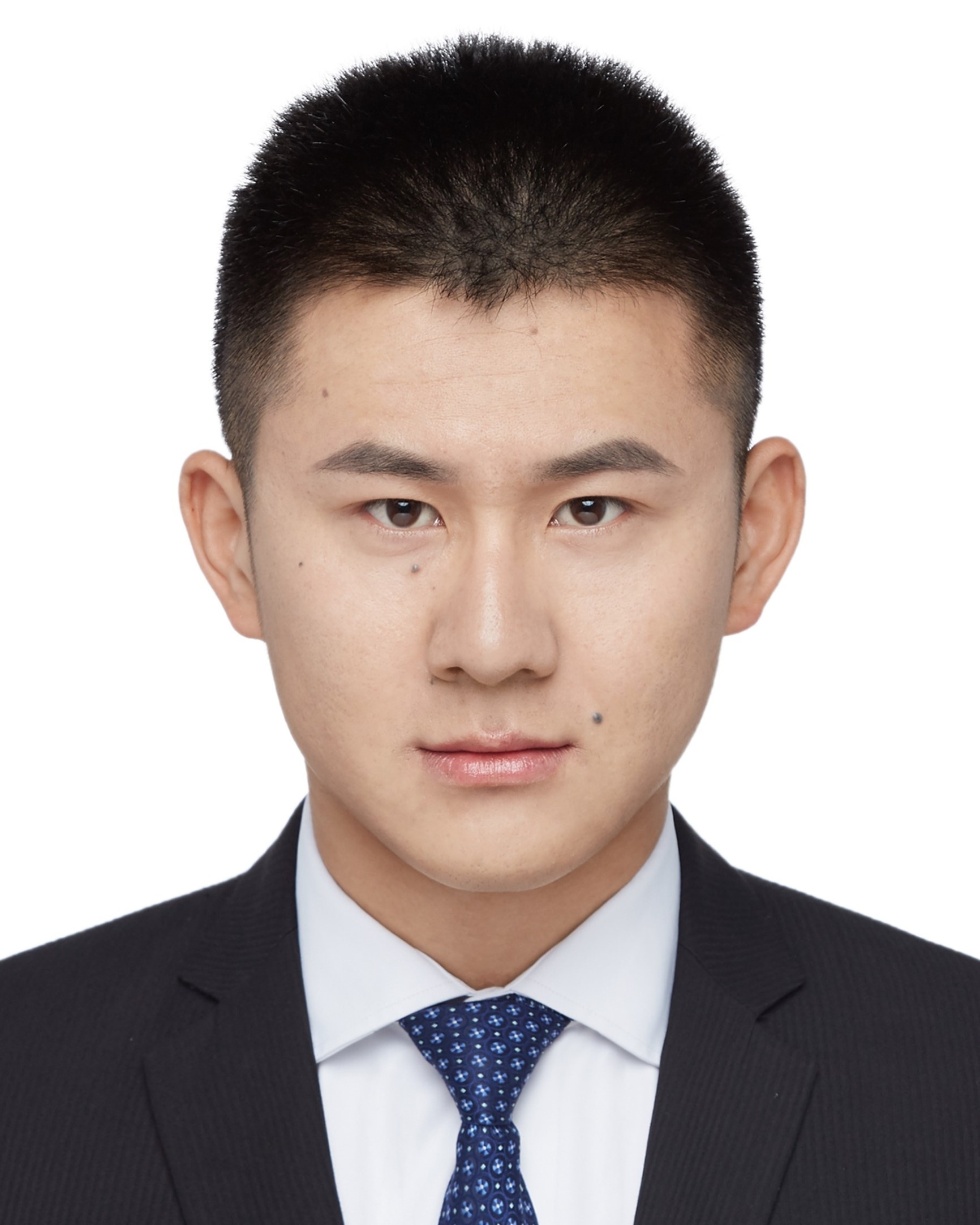}}]{Guihong Li}
(Student Member, IEEE) received the B.S degree from the Beijing University of Posts and Telecommunications, Bejing, China, in 2018. He is currently pursuing his Ph.D. in Electrical and Computer Engineering at The University of Texas at Austin, USA. His research interest includes Neural Architecture Search, hardware-software co-design for EdgeAI system optimization. He received many awards including a best paper nomination from ISWC 2022.
\end{IEEEbiography}

\begin{IEEEbiography}[{\includegraphics[width=1in,height=1.25in,clip,keepaspectratio]{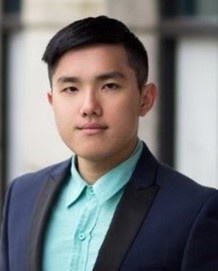}}]{Duc Hoang} recevied a Bachelor's degree in ECE from the University of Washington. He is currently pursuing a Ph.D. in the same field at the University of Texas at Austin, focusing on Graph Neural Networks, Neural Architectural Search, and Network Pruning. 
\end{IEEEbiography}


\begin{IEEEbiography}[{\includegraphics[width=1in,height=1.25in,clip,keepaspectratio]{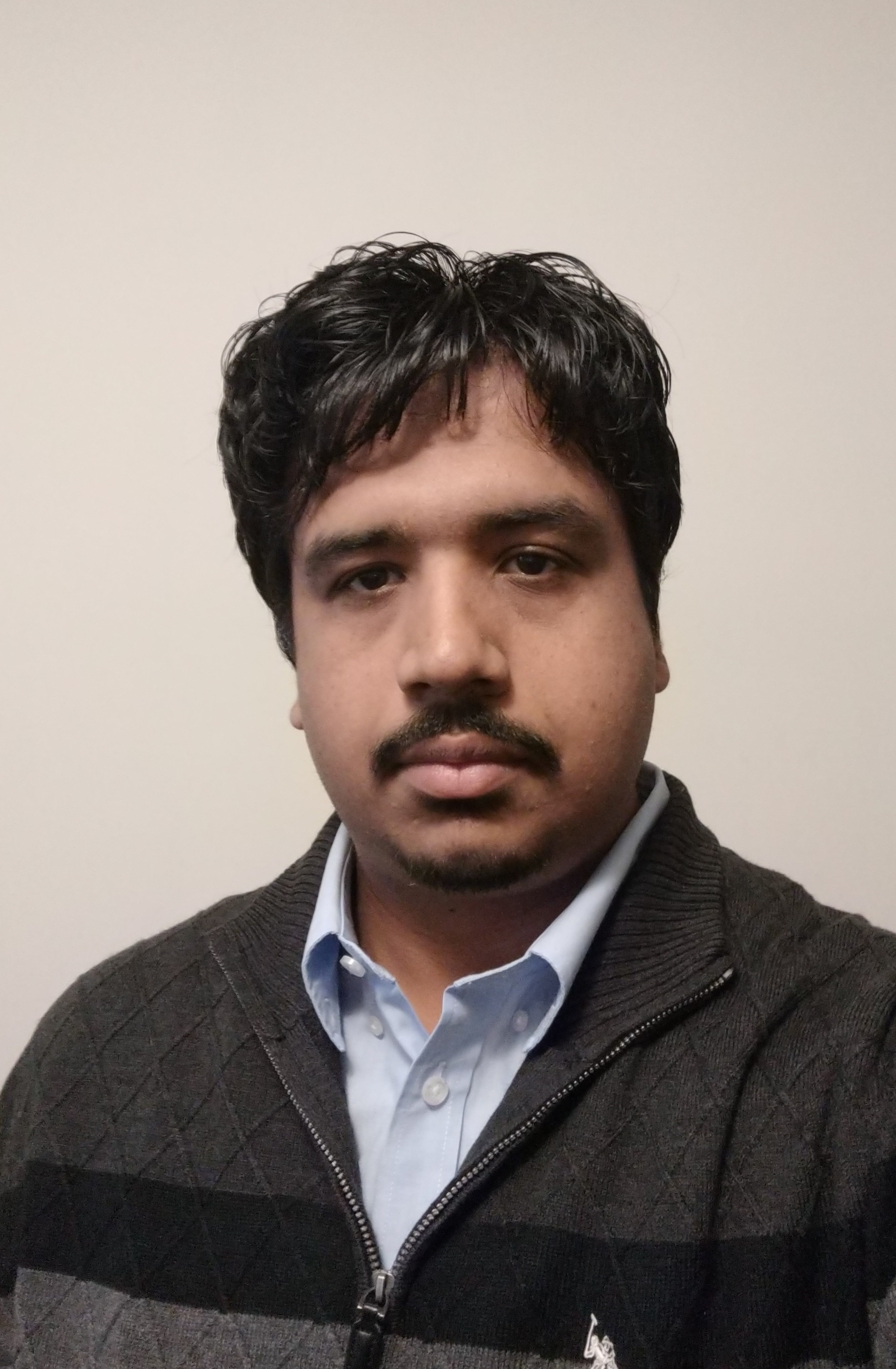}}]{Kartikeya Bhardwaj} is a Senior Machine Learning Researcher at Qualcomm AI Research. Previously, he was a Senior Machine Learning Engineer at Arm, Inc. He completed his PhD in Electrical and Computer Engineering from Carnegie Mellon University in 2019. His research interests are in
 the field of hardware-aware deep learning, computer vision, and network science. His work has been published in top conferences including CVPR, ICLR, MLSys, ECML, DAC, DATE, etc.
\end{IEEEbiography}

\begin{IEEEbiography}[{\includegraphics[width=1in,height=1.25in,clip,keepaspectratio]{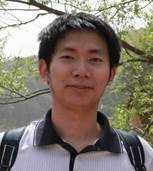}}]{Ming Lin}
is a Senior Applied Scientist at Amazon.com LCC. His research interests include Mathematical Foundation of Deep Learning and Statistical Machine Learning, with their applications in deep learning acceleration, computer vision and mobile AI. He worked as a postdoctoral researcher in the School of Computer Science at Carnegie Mellon University from July 2014 to Sep 2015. He received his Ph.D. degree in computer science from Tsinghua University in 2014. During his Ph.D. study, he had been a visiting scholar in Michigan State University and in CMU from Dec 2013 to July 2014.\end{IEEEbiography}

\begin{IEEEbiography}[{\includegraphics[width=1in,height=1.25in,clip,keepaspectratio]{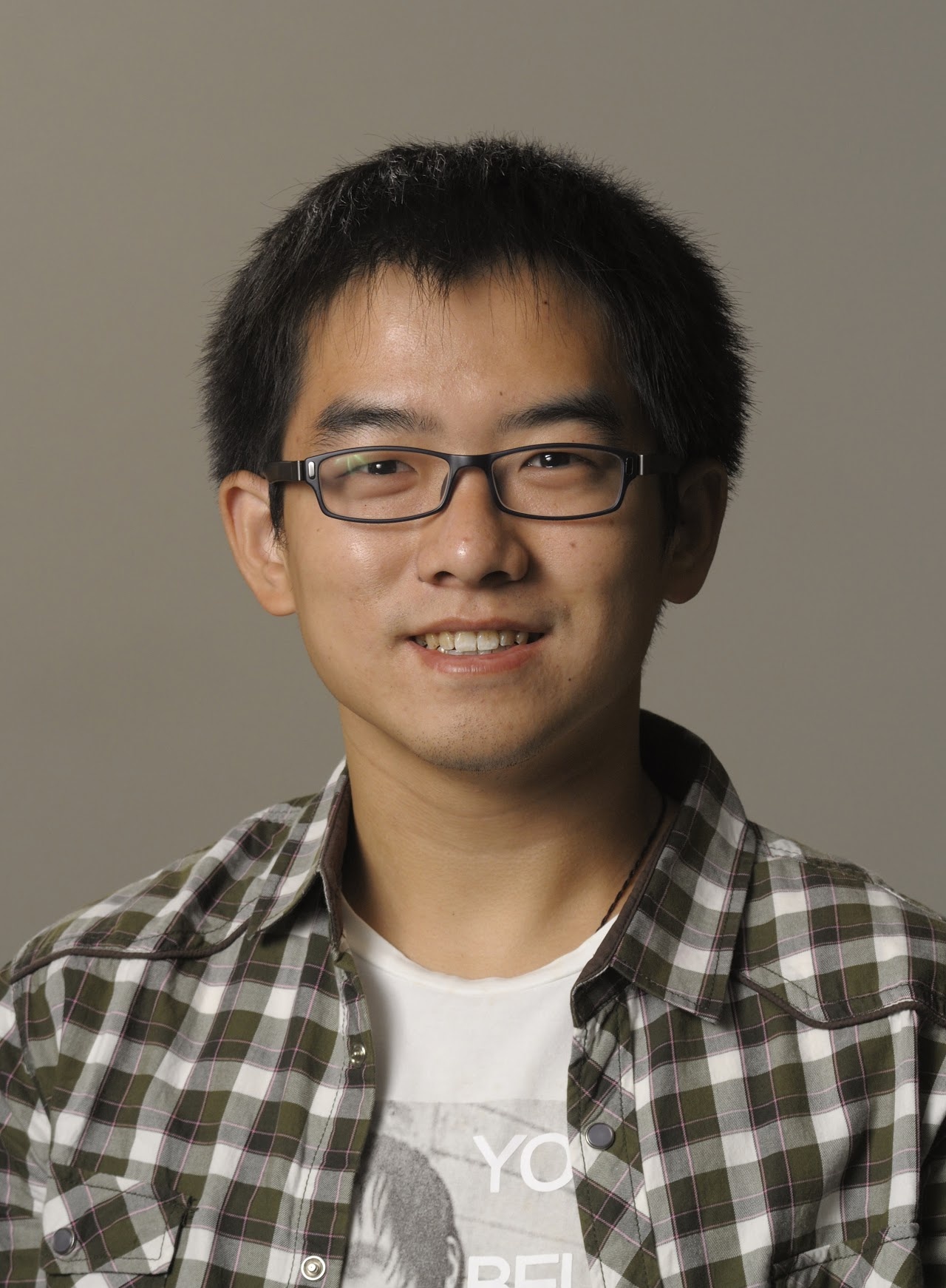}}]{Zhangyang Wang} is currently the  Temple Foundation Endowed Associate Professor \#7 of ECE at UT Austin. He received his Ph.D. in ECE from UIUC in 2016, and his B.E. in EEIS from USTC in 2012. Prof. Wang is broadly interested in the fields of machine learning, computer vision, optimization, and their interdisciplinary applications. His latest interests focus on the role of low dimensionality in deep learning.
\end{IEEEbiography}

\begin{IEEEbiography}[{\includegraphics[width=1in,height=1.25in,clip,keepaspectratio]{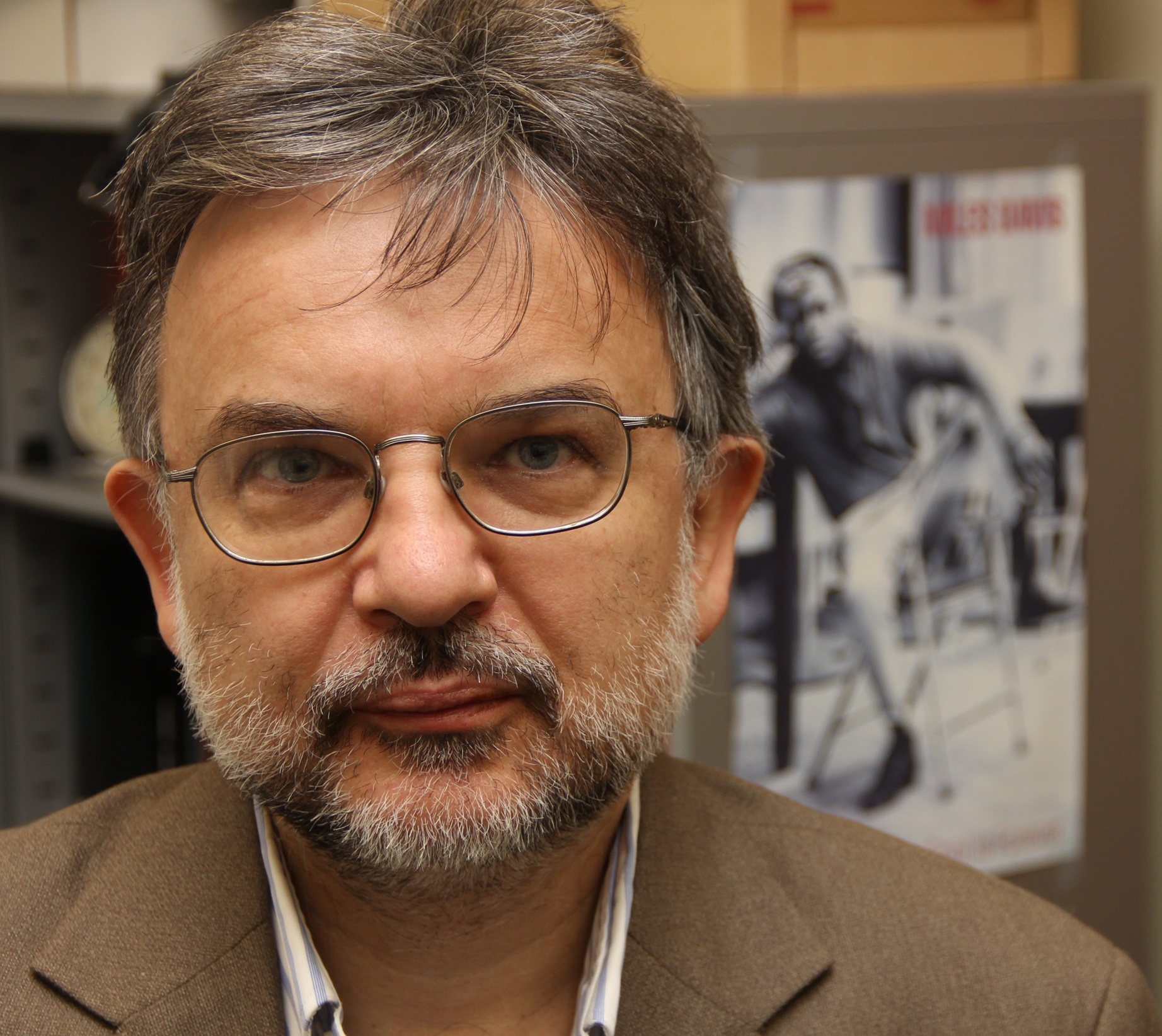}}]{Radu Marculescu}
is the Laura Jennings Turner Chair in Engineering and Professor in the Electrical and Computer Engineering department at The University of Texas at Austin. He received his Ph.D. in Electrical Engineering from the University of Southern California in 1998. Radu’s current research focuses on developing ML/AI methods and tools for modeling and optimization of embedded systems, cyber-physical systems, and social networks.
\end{IEEEbiography}




\end{document}